\documentclass[journal]{./sty/IEEEtran}

\usepackage{algorithm}
\usepackage{algorithmic}
\usepackage{graphicx}
\usepackage{subfigure}
\usepackage{multirow}
\usepackage{cite}
\usepackage{subfigure}
\usepackage{amsmath}
\usepackage{flushend}
\usepackage{array}
\usepackage{url}
\usepackage{amssymb}

\begin{document}

\title{Exploring Human Mobility for Multi-pattern Passenger Prediction: A Graph Learning Framework}

\author{Xiangjie Kong,~\IEEEmembership{Senior Member,~IEEE},
	Kailai Wang, Mingliang Hou, Feng Xia,~\IEEEmembership{Senior Member,~IEEE}, 
	\\ Gour Karmakar,~\IEEEmembership{Member,~IEEE}, and Jianxin Li
	
	\thanks{This work was supported in part by the National Natural Science Foundation of China under Grant 62072409, in part by the Zhejiang Provincial Natural Science Foundation under Grant LR21F020003, and in part by the Fundamental Research Funds for the Provincial Universities of Zhejiang under Grant RF-B2020001. \emph{(Corresponding author: Feng Xia.)}}
	
	\thanks{X. Kong is with the School of Software, Dalian University of Technology, Dalian 116620, China, and also with the College of Computer Science and Technology, Zhejiang University of Technology, Hangzhou 310023, China (e-mail: xjkong@ieee.org).}
	\thanks{K. Wang and M. Hou are with the School of Software, Dalian University of Technology, Dalian 116620, China (e-mail: kailai.w@outlook.com; teemohold@outlook.com).} 
	\thanks{F. Xia is with the School of Engineering, IT and Physical Sciences, Federation University Australia, Ballarat 3353, Australia (e-mail: f.xia@ieee.org).}
	\thanks{G. Karmakar is with the School of Engineering, IT and Physical Sciences, Federation University Australia, Churchill 3842, Australia (e-mail: gour.karmakar@federation.edu.au).}
	\thanks{J. Li is with the School of IT, Deakin University, Melbourne 3125, Australia (e-mail: jianxin.li@deakin.edu.au).}
}

\markboth{IEEE TRANSACTIONS ON INTELLIGENT TRANSPORTATION SYSTEMS}{}

\maketitle

\begin{abstract}
  Traffic flow prediction is an integral part of an intelligent transportation system and thus fundamental for various traffic-related applications. Buses are an indispensable way of moving for urban residents with fixed routes and schedules, which leads to latent travel regularity. However, human mobility patterns, specifically the complex relationships between bus passengers, are deeply hidden in this fixed mobility mode. Although many models exist to predict traffic flow, human mobility patterns have not been well explored in this regard. To reduce this research gap and learn human mobility knowledge from this fixed travel behaviors, we propose a multi-pattern passenger flow prediction framework, MPGCN, based on Graph Convolutional Network (GCN). Firstly, we construct a novel sharing-stop network to model relationships between passengers based on bus record data. Then, we employ GCN to extract features from the graph by learning useful topology information and introduce a deep clustering method to recognize mobility patterns hidden in bus passengers. Furthermore, to fully utilize Spatio-temporal information, we propose GCN2Flow to predict passenger flow based on various mobility patterns. To the best of our knowledge, this paper is the first work to adopt a multi-pattern approach to predict the bus passenger flow from graph learning. We design a case study for optimizing routes. Extensive experiments upon a real-world bus dataset demonstrate that MPGCN has potential efficacy in passenger flow prediction and route optimization.
  
\end{abstract}


\IEEEpeerreviewmaketitle

\section{Introduction}
\IEEEPARstart{S}{mart} cities have been gradually formed by information and communication technologies, including the Internet of Things (IoT)~\cite{Sisinni2018Iot}, cloud computing~\cite{Rahimi2018mobile}, and edge computing~\cite{kong2020mobileEdge}. One important application scenario of the smart city is the Intelligent Transportation System (ITS) to improve the public services and gain the solutions to problems in urban transportation such as traffic jams, traffic accidents, parking chaos, route planning~\cite{Kong2018subway}, and resource allocation~\cite{Ruan2020dynamic}. The above problems are closely related to traffic flow and its prediction. Furthermore, the smart city industry also plays an important role in Big Data and generates various Spatio-temporal data styles~\cite{Gowtham2018stdata, Du2019city, Wang2020deep} including GPS, sensors, social media, and traffic cards. Driven these urban Spatio-temporal Big Data, the main challenge a smart city faces can be summarized in two aspects: (1) how to deal with and analyze large but redundant Spatio-temporal data, and (2) how to improve human mobility and optimize travel.

\par Public transportation accounts for a large proportion of urban transportation. Taking Beijing as an example, buses produced 1.7 billion vehicle kilometers traveled and transported 4.9 billion passengers in 2011 alone~\cite{Zhou2014commuting}. The behavior that people are encouraged to take buses is beneficial to the city's sustainable development owning to low-carbon and green mode of buses. Therefore, the operation management of public transportation directly affects the traffic circumstance of the city, which the government has always valued. Many policies have also been adopted to try to improve public transportation, such as preferential bus fares, bus lanes, additional stops, routes, and bus running time~\cite{pili2019evaluating}. 

\par
\begin{figure}
    \centering
    \includegraphics[width=0.3\textwidth]{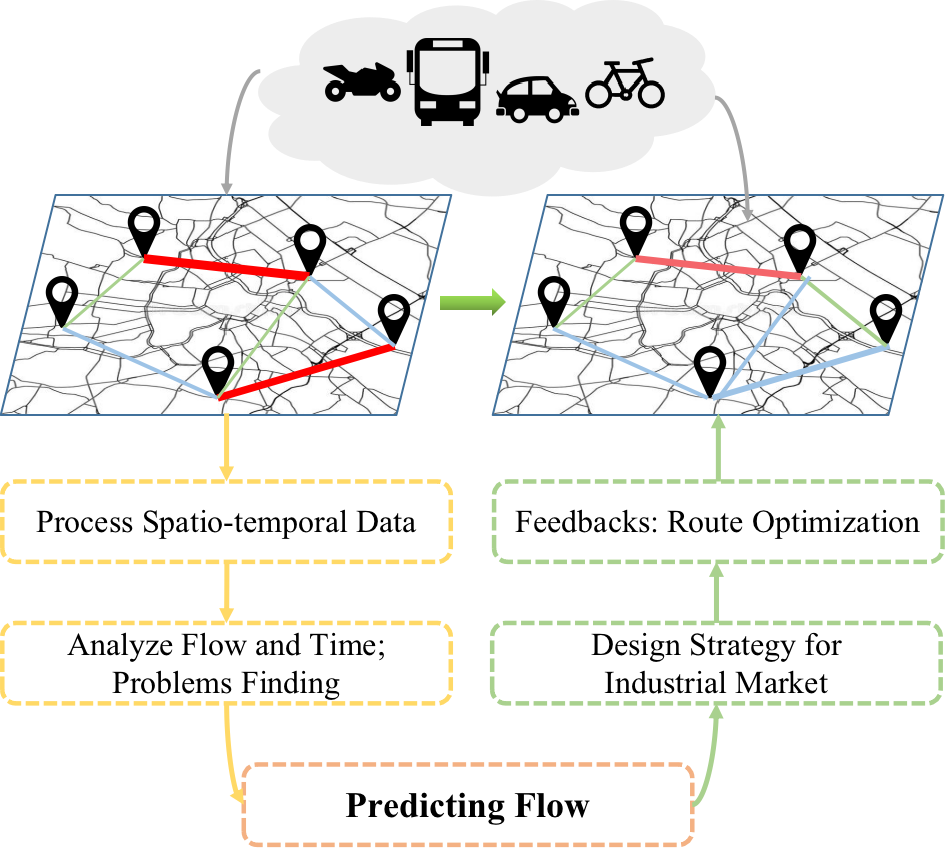}
    \caption{Application and process of traffic flow prediction. }
    \label{fig:application}
\end{figure}

 However, emerging traffic problems like severe traffic congestion and the unreasonable allocation of resources have urged researchers to study~\cite{Mourad2019survey}. Consequently, the new services and requirements are urgent to improve bus travel and ride experiences, where the passenger flow prediction becomes critical. As shown in \figurename{ \ref{fig:application}}, the existing potential problems are solved through processing and analyzing Spatio-temporal data. One of the most effective solutions is route optimization, which is a complex and challenging task, although valuable to the related industry in sustainable transportation systems. It is not difficult to find that traffic flow prediction is essential in the whole process of route optimization. If we predict the traffic flow accurately, we can respond in time to avoid traffic jams and keep roads smooth. 


\par Due to this demand for flow prediction, plenty of work has contributed to traffic prediction for a long time. In general, some traffic flow prediction approaches are built on traditional mathematical statistics~\cite{Gan2014Seasonal}, such as AR, ARMA, and ARIMA. On the other hand, as a result of the limitations of traditional models and the excellent performance of deep learning in prediction tasks, the deep learning-based methods have been evolved, like DNN~\cite{Liu2017passenger}, DBN~\cite{Bai2017multi}, LSTM~\cite{Du2020LSTM}, and GAN~\cite{zhang2019trafficgan}. However, the methods mentioned above only consider the numerical traffic flow based on the statistics of Spatio-temporal data but neglect the existence of human mobility behavior, which refers to travel habits and plays a decisive role in the change of traffic flow. The issue can lead to a lack of definite identification and differentiation of traffic flow. Therefore, existing works do not identify mobility behaviors through the relationship between peoples, which results in a deviation for the final prediction performance of different groups. Experientially, people in a similar group have similar mobility behaviors. For example, the actuality that most commuters work from 9 am to 5 pm may mean that they must travel at least once in the morning and evening and pass regular bus stops. Therefore, our studies defines a passenger mobility pattern as a group of people with similar travel routes. Hypothetically, the total flow consists of two flows: (i) steady flow having from most people with permanent jobs and residence, and (ii) uncertain flow generated from travel, entertainment, and so on. However, few studies considered passenger mobility pattern analysis to predict traffic flow as far as we know. 

\par In recent years, Graph Neural Network (GNN), especially Graph Convolutional Network (GCN), has effective performance in extracting the features and relationships of a topological graph. GNN not only represents explicitly the nodes of the graph in the low-dimensional vector space (also called embedding), but retains essential attributes~\cite{Xu2019how}. Ordinarily, the embeddings of nodes can be used in various downstream tasks, including clustering~\cite{shen2021gcncdd}, classification, and prediction~\cite{han2020congestion}. Different from the previous methods treating Spatio-temporal information of the trajectory data as the main feature~\cite{pan2019stnn}, based on the passenger bus record data, we try to define and construct an interpretable graph structure-based network that can be applied in GCN based model to explore passenger mobility patterns. Besides, the adjacent bus stops with a solid Spatio-temporal relationship can improve the accuracy of traffic flow prediction in the mass transit network. It is important to note that our studies focus on the prediction, so we do not consider the fare evasion~\cite{barabino2020fare} and certain anomalies~\cite{mcleod2007estimating, barabino2017time}, which do not affect the evaluation of the predictive model.

\par In this paper, we introduce a framework, namely MPGCN, with three stages to predict passenger flow for the first time. We first obtain related information of bus stops based on the bus record data, which are used in the analysis of passenger mobility patterns. Secondly, we construct a sharing-stop network of passengers, including stop matching and weight assignment of graph edges. The sharing-stop network is utilized in graph deep clustering with GCN to explore mobility patterns. Furthermore, to verify their diversity, we execute the statistical analysis of each mobility pattern by describing heavy-tailed distributions of the number of bus stops the passengers passed. Then, considering the spatial correlation of bus route network and temporal correlation of traffic flow, we propose GCN2Flow combining Spatio-temporal information to separately predict passenger flow of different patterns, where the predictive flows for all passenger patterns are fused in the final stage to obtain the prediction result. Finally, we design a case study for optimizing routes, where we select optimal routes from candidates of passengers and set the passenger diversion and experience as the main optimization objective based on previous prediction results.

\par The main contributions of this work are summarized as follows:

\begin{itemize}
\item We develop a novel prediction framework, namely MPGCN, which integrates the passenger mobility patterns into passenger flow prediction task to enhance accuracy. We define a passenger mobility pattern as a group of people with similar travel times or similar travel routes. Our MPGCN includes three stages to achieve the prediction: (i) pre-processing the bus record data, (ii) recognizing passenger mobility patterns, and (iii) predicting passenger flow by proposed GCN2Flow combined with Spatio-temporal information. Besides, we design constrained planning as a case study for optimizing routes and thus improving passenger diversion and experience based on the prediction results.	

\item We present a sharing-stop network, where the relationship between passengers is established, and explore the passenger mobility patterns in the sharing-stop network based on deep clustering with GCN. Through statistical analysis and data fitness, we suggest the reasonability and interpretability of the network, as well as the significant laws among different mobility patterns.

\item We conduct a series of experiments, including the analysis of passenger mobility patterns and the comparison with different prediction algorithms with or without passenger mobility patterns. The predictive evaluation demonstrates that our framework has better performance and substantially improves passenger flow prediction. Besides, Prediction is accurate enough to be used for downstream tasks such as route optimization.
\end{itemize}

\par The remainder of this paper is structured as follows. In Section 2, we briefly review related works. Section 3 mainly illustrates our proposed model and framework (MPGCN) in detail. Data description and analysis of the experimental results are given in Section 4. Finally, we present discussions, conclusion and future work in Section 5 and 6.

\section{Related Work}
\par In this section, we review the existing works closely related to the research project presented in this paper covering three fields: traffic flow prediction, human mobility pattern, and graph convolutional network. 

\subsection{Traffic Flow Prediction}
\par Traffic flow prediction has many cutting edge applications, such as road network planning, congestion prevention, and accident detection. Considering the technical approach applied in prediction, traffic flow prediction models can be roughly into three categories: (i) traditional mathematical-statistical parametric, (ii) non-parametric regression, and (iii) artificial neural network (ANN) models.

\par The mathematical-statistical parametric models mainly examine the time series that have a periodic change rule in the urban road traffic, like traffic peak in the day and night time. These models include autoregressive moving average (ARMA)~\cite{Sadek2004multiscale}, and autoregressive integrated moving average (ARIMA)~\cite{Kumar2015short}.

Kumar et al.~\cite{Kumar2015short} utilized the seasonal ARIMA model to design a prediction scheme using only limited input data. In this model, the issues associated with huge time-series data like availability, computation, storage, and maintenance are considered, and the last three days’ flow observations were used as input for predicting the next day’s flow. Support Vector Regression (SVR)~\cite{Ge2019forecasting} and Nearest Neighbor Regression~\cite{Dell2015time}  are the most popular nonparametric regression models.
With the complexity and diversity increasing in the road network, there is a demand for more accurate traffic prediction. Because of the efficacy of ANN for various prediction tasks in complex and diverse scenarios, it attracts attention to traffic flow prediction. Liu et al.~\cite{Liu2017passenger} proposed a novel passenger flow prediction model using a hybrid of deep network of unsupervised stacked autoencoders (SAE), and supervised deep neural network (DNN). Besides, Yu et al.~\cite{yu2018spatio} introduced Spatio-temporal GCN (STGCN) model to forecast traffic using graph GNN and Gate CNN for extracting spatial and temporal features, respectively. More recent works such as T-GCN~\cite{zhao2020tgcn}, and TGC-LSTM~\cite{cui2020traffic} further utilized GCN to the extraction of spatial and temporal correlations to improve prediction. However, all the models presented in this section only focus on mining the information in the raw data while ignoring the potential mobility patterns of passengers. For addressing this research issue, for the first time, this paper aims to discover passenger mobility behaviors and rules from the bus record data and then utilize them for traffic flow prediction. Since passenger mobility behaviors and rules are to be exploited in this paper to predict passenger flow, human mobility pattern is presented in the next section.

\subsection{Human Mobility Pattern}
\par The fluctuation of traffic flow naturally depends on the human mobility and travel. Consequently, the analysis of human mobility patterns is of paramount importance for traffic prediction. The current methods are distinguished primarily by the type of datasets. For example, two dataset types: (i) unconscious mobility data (e.g., sensors data, or card records) and (ii) active sharing mobility data (e.g., traditional diary activity surveys or social location sharing). Using the former dataset type, Yan et al.~\cite{Yan2014universal} presented a model to capture the underlying driving force accounting for human mobility patterns based on GPS and mobile phone data. Besides, Qi et al.~\cite{qi2019analysis} designed a multi-step methodology to extract mobility patterns from smart card data and points of interest data. Nitti et al.~\cite{nitti2020iabacus} presented a Wi-Fi-based Automatic Bus pAssenger CoUnting System (iABACUS), which did not depends on passengers card records and can track passengers to analyze urban mobility. In \cite{Xia2018exploring}, authors integrated taxi and subway data to compute the human mobility network, and discover the human mobility patterns in terms of trip displacement, duration, and interval. On the other hand, people are willing to share their activities containing location information because of the convenience and popularity of the social networks, like Weibo and Twitter. Utilizing the latter, Comito et al.~\cite{Comito2016mining} developed a methodology to discover people, community behavior and travel routes from geo-tagged posts and tweets. Nevertheless, none of these research works has leveraged human mobility patterns to predict traffic flow. Therefore, exploring human mobility patterns from passenger bus data is one of the main aims of the research project presented in this paper. Since we decide to leverage graph convolutional network for the traffic flow prediction, the next section presents its overview.

\begin{figure}[htb]
	\centering
	\includegraphics[width=0.45\textwidth]{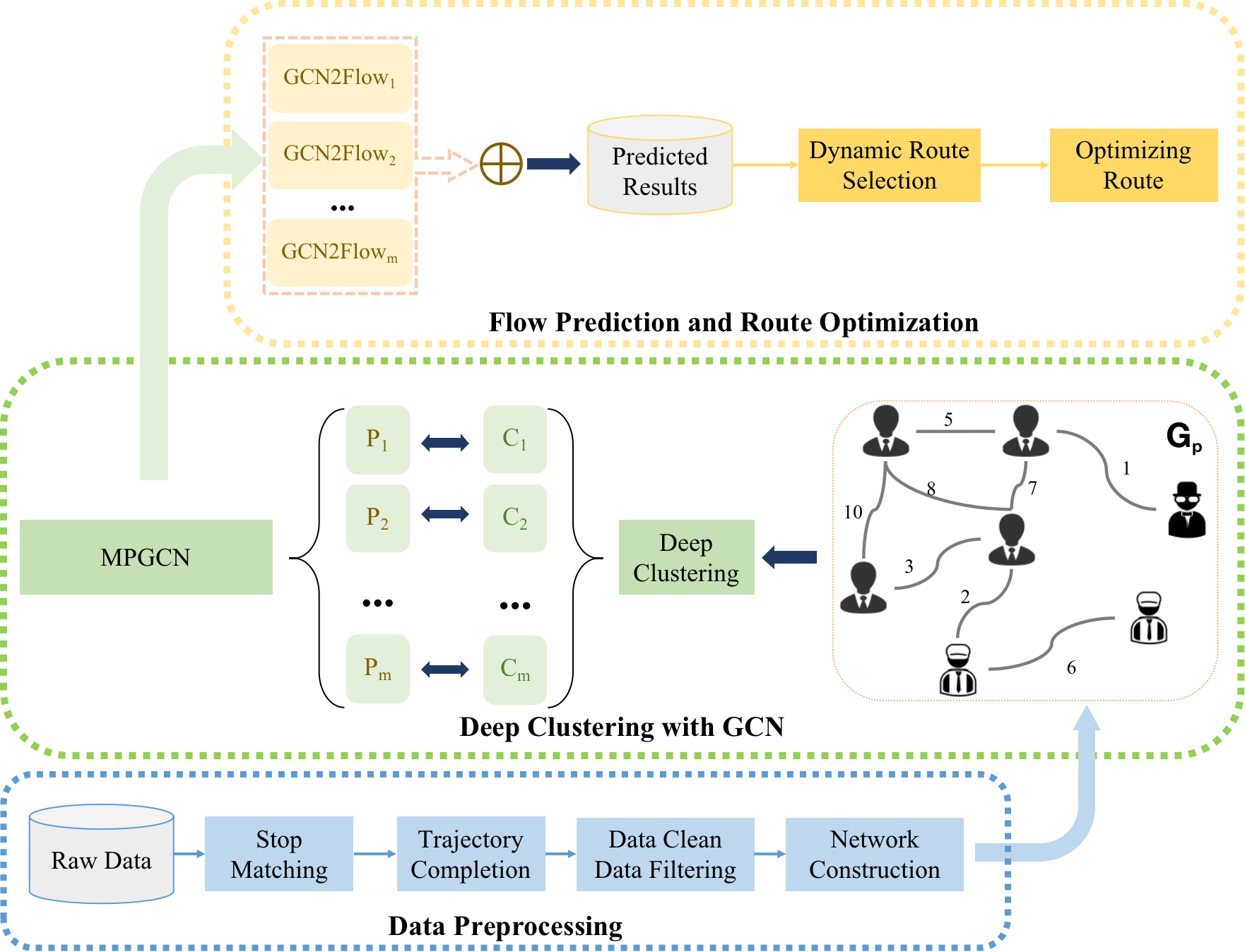}
	\caption{Framework of the model, MPGCN, where $C_m$ and $P_m$ represent cluster and corresponding pattern of passenger mobility.}
	\label{fig:framework}
\end{figure}

\subsection{Graph Convolutional Network}
\par With the development of graph learning, there is an extension of CNNs in the graph (network) as an extensive data structure, like embedding of nodes or subgraphs into vector spaces. The first convolutional operation on graphs is presented in ~\cite{Bruna2014spectral}.  However, it has been evolved over time for its effectiveness in representing graph~\cite{kipf2016variational} and its numerous application domains, such as node clustering~\cite{Bo2020Structural}, classification~\cite{kipf2017semi}, prediction~\cite{chai2018bike}, and so on. Besides, to further utilize neighbours’ information, GAT~\cite{velivckovic2017graph} combined multi-head self-attention mechanism to calculate attention scores of different neighbours. In this paper, unlike others, a major challenge is to establish an explainable network from bus record data, which can enable us to discover the relationship between passengers. Besides, we also need to extract spatial features of geographical information in the stop network to improve passenger flow prediction.

\section{Design of Framework}
\par This section provides the details of the theoretical underpinning of our proposed network and techniques used in our proposed Multi-Pattern GCN based passenger flow prediction, namely MPGCN framework as shown in \figurename{ \ref{fig:framework}}. 

\subsection{Network  construction: Sharing-Stop Network}
\par For exploring human mobility patterns, a sharing-stop network is defined as a weighted undirected graph $\mathcal{G}_p=\left\lbrace \mathcal{V}_p, \mathcal{E}_p, \mathcal{A}_p \right\rbrace $, where $\mathcal{V}_p$ is the set of passenger nodes and $\mathcal{E}_p$ is the set of edges. $\mathcal{A}_p \in \mathbb{R}^{\left|\mathcal{V}_p\right| \times \left|\mathcal{V}_p\right|}$ is the weighted matrix with each element $a^p_{ij} \ge 0$. Specifically, the edge between passengers $i$ and $j$ denotes their existing relationship of sharing stops, and weight $a^p_{ij}$ indicates the count of the occurrences of their sharing stops. The pseudocode of the construction of the sharing-stop network is shown in Algorithm \ref{alg:Construction}. A simple example of this sharing-stop network is shown in the middle part of \figurename{ \ref{fig:framework}} based on the empirical assumption that there are similar records between passengers in the same pattern meaning that for the same mobility pattern, passengers have a similar number of edges and the similar value of their weights in the sharing-stop network.
\begin{algorithm}
	\caption{The construction of sharing-stop network}
	\label{alg:Construction}
	\begin{algorithmic}[1]
		\REQUIRE $P2S$: the list of passengers to stops;
		\ENSURE $G_p$, sharing-stop network;
		\STATE Initialize the graph structure $G_p$;
		\FOR{$p_i, p_j$ in $P2S.keys$}
			\STATE $s_i, s_j \gets P2S[p_i], P2S[p_j]$; $\backslash\backslash$ obtain stops set of records;
			\STATE $s\_s \gets s_i \cap s_j$; $\backslash\backslash$ obtain all sharing-stops;
			\IF{$s\_s$ is empty}
				\STATE continue;
			\ENDIF
			\STATE Initialize $a^p_{ij}=0$ in $G_p$; $\backslash\backslash$ add an edge;
			\FOR{$s$ in $s\_s$}
				\STATE $c_i, c_j \gets s_i.count(s), s_j.count(s)$; $\backslash\backslash$ counting;
				\STATE $G_p.a^p_{ij} += min(c_i, c_j)$; $\backslash\backslash$ update the weight;
			\ENDFOR
		\ENDFOR
		\RETURN $G_p$;
	\end{algorithmic}
\end{algorithm}

\subsection{Deep Clustering with GCN}

\begin{figure}[htb]
	\centering
	\includegraphics[width=0.45\textwidth]{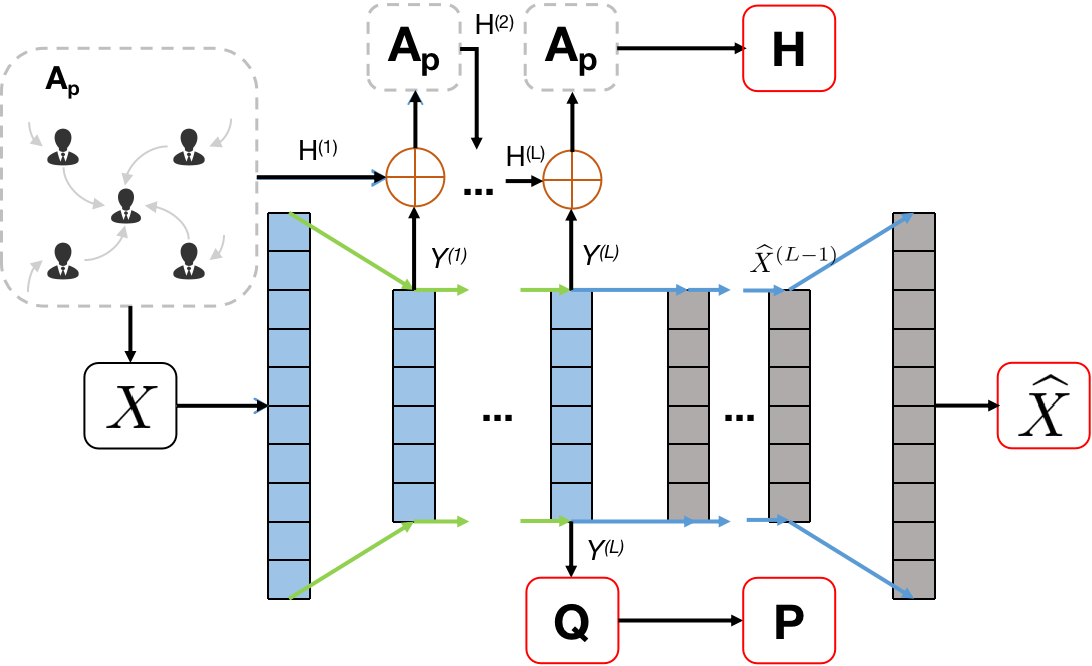}
	\caption{Computational process of deep clustering with GCN.}
	\label{fig:deepClustering}
\end{figure}

\par After constructing the graph, inspired by~\cite{Bo2020Structural}, an unsupervised deep clustering with GCN is used to mine potential passenger mobility patterns as shown in \figurename{ \ref{fig:deepClustering}}. First, unsupervised representation method, the basic autoencoder (AE), is employed to learn the representation of passenger nodes, that is a mapping function $ \Phi:v\in \mathcal{V}_p \mapsto \mathbb{R}^{d} $, where $ d\ll \left|\mathcal{V}_p\right|$. We assume that there are $L$ layers in the encoder and decoder parts, which are symmetrical. Therefore, the representation of the $l$-th layer, $Y^{(l)}$ in the encoder and $\hat{X}^{(l)}$ in the decoder, can be obtained as follows:
\begin{equation}
	\left\{
	\begin{aligned}
	&Y^{(l)}=\phi\left(W_e^{(l)}Y^{(l-1)}+b_e^{(l)}\right),\\
	&\hat{X}^{(l)}=\phi\left(W_d^{(l)}\hat{X}^{(l-1)}+b_d^{(l)}\right),\\
	\end{aligned}
	\right.
\end{equation}
where $W$ and $b$ are the weight matrix and bias, respectively, and $\phi$ is the activation function, such as Relu or Sigmoid functions. Besides, the input of the encoder, $Y^{(0)}$ is X that is initial feature matrix obtained by previous sharing-stop network, and the output of the decoder is the reconstruction of $\hat{X}=\hat{X}^{(L)}$. Hence, the loss function of the entire AE is as follows:
\begin{equation}
\mathcal{L}_1=\dfrac{1}{2|V|}\left\| X-\hat{X}\right\|^2,
\end{equation}
where $||\cdot||$ denotes the Euclidean distance between two representations matrix.

\par On the other hand, we integrate these representations into GCN that can learn them by combining the relationship between passenger nodes. In this part, the convolutional operation of the $l$-th layer can be defined by:
\begin{equation}
H^{(l)}=\phi\left( \tilde{D}^{-\frac{1}{2}}\tilde{A}_p\tilde{D}^{-\frac{1}{2}}\tilde{H}^{(l-1)}W^{(l-1)} \right),
\end{equation}
where $\tilde{A}_p=\mathcal{A}_p+I$, and $I$ is an identity matrix. $\tilde{D}_{ii}=\sum_{j}\tilde{a}^p_{ij}$ is the degree of node $i$ in adjacent matrix $A_p$, and $W$ is the weight matrix of parameters. Specially, the input of $l$-th layer in GCN, $\tilde{H}^{(l-1)}$, combines the representations from the initial GCN and AE:
\begin{equation} \label{eq:GCN-AE}
\tilde{H}^{(l-1)}=\alpha H^{(l-1)}+(1-\alpha)Y^{(l-1)},
\end{equation}
Eq. (\ref{eq:GCN-AE}) joins GCN with AE. In this case, we uniformly set $\alpha=0.5$. The final representation of the last layer can be mapped as a multiple classification probability with softmax function:
\begin{equation}
H=softmax\left(\tilde{D}^{-\frac{1}{2}}\tilde{A}_p\tilde{D}^{-\frac{1}{2}}H^{(L)}W^{(L)}\right),
\end{equation}
where $H$ can be regarded as the probability distribution, and $h_{ij}\in H$ denotes the probability of node $i$ in cluster $j$.

\par For being more suitable for deep clustering, a dual self-supervised method is employed to combine clustering information with the representation learned previously. With t-distribution to measure the similarity, the probability of node $i$ in cluster $j$ is as follows:
\begin{equation}
\label{eq_cluster}
q_{ij}=\dfrac{\left(1+\left\| y_i-\mu_j\right\|^2/n \right)^{-(n+1)/2}}{\sum_{j}\left(1+\left\| y_i-\mu_j\right\|^2/n \right)^{-(n+1)/2}},
\end{equation}
where $y_i$ is from the $Y^{(L)}$, $\mu_j$ is the cluster center vector initialized by the K-means clustering, and $n$ is the degree of freedom of t-distribution. Therefore, we obtain the clustering probability distribution $Q=\left\lbrace q_{ij}\right\rbrace $. Besides, the target distribution, $P$, can be computed and normalized as follows:
\begin{equation}
p_{ij}=\dfrac{q^2_{ij}/\sum_{i}q^2_{ij}}{\sum_{k}(q^2_{ik}/\sum_{i}q^2_{ik})}.
\end{equation}
Then, we use the KL divergence as part of the loss function, that is $\mathcal{L}_2$ is the KL divergence between $P$ and $Q$ distributions, and $\mathcal{L}_3$ is between $P$ and $H$. In the end, the overall loss function is defined by:
\begin{equation}
\mathcal{L}=\theta_1\mathcal{L}_1+\theta_2\mathcal{L}_2+\theta_3\mathcal{L}_3.
\end{equation}
where $\theta$ is the hyper-parameters, $\mathcal{L}_2=KL(P||Q)$, and $\mathcal{L}_3=KL(P||H)$. The final cluster label of node $i$ is determined considering the maximum value of $h_{ij}$ from the probability distribution $H$.

\subsection{Multi-Pattern GCN Based Passenger Flow Prediction}
\par The flow prediction components include identifying passenger patterns, training neural network model with GCN, and predicting passenger flow. From the previous results of clustering, passenger nodes under the same cluster reflect that they have similar mobility rules. Therefore, we divide the passengers into several mobility pattern groups utilizing the clustering results, which is also patterns exploration. Besides, we design a statistical task to discover the potential law by fitting several possible heavy-tailed distribution~\cite{Xia2018exploring} shown in Section \ref{experiments}.

\begin{figure}[htb]
	\centering
	\includegraphics[width=0.3\textwidth]{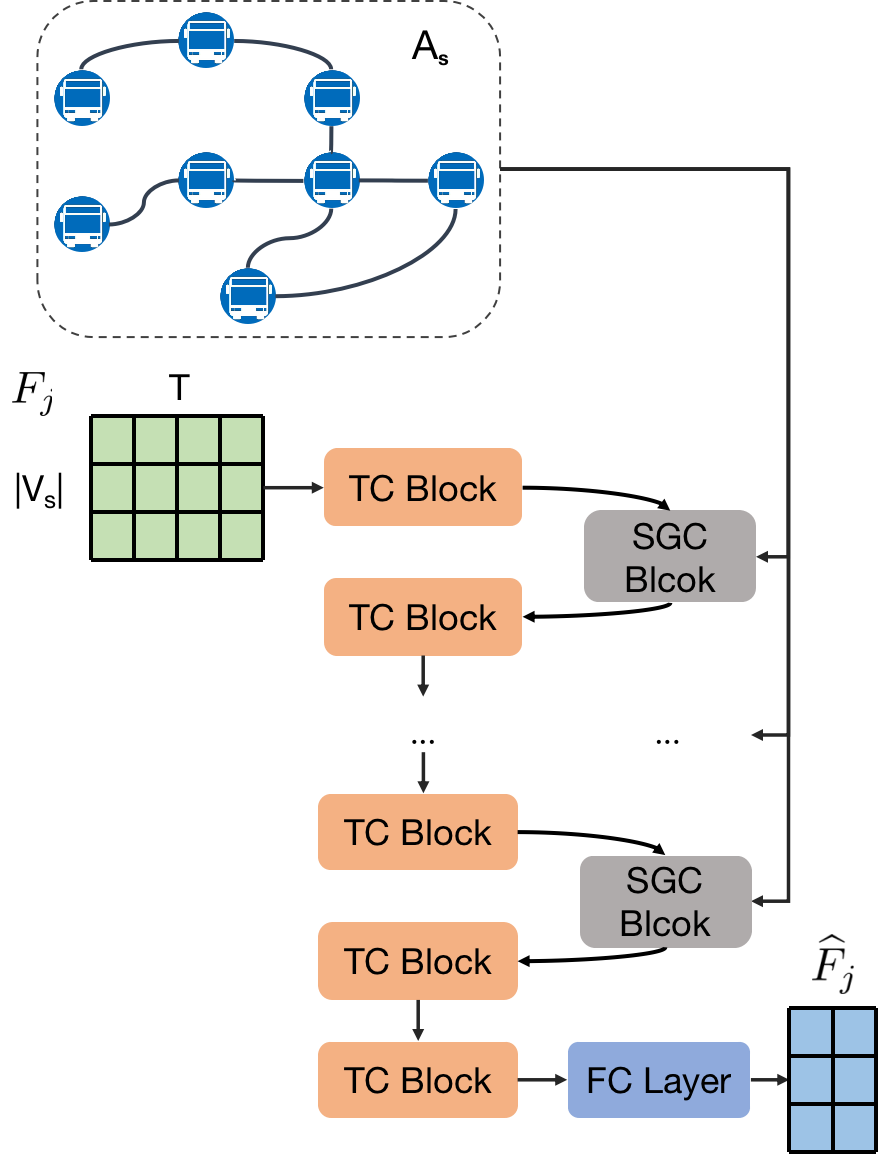}
	\caption{The architecture of GCN2Flow, which takes spatial bus route network ($A_s$), and time series flows for the passengers in pattern $j$ ($F_j$) as input of TC block and SGC block.}
	\label{fig:GCN2Flow}
\end{figure}

\par As shown in \figurename{ \ref{fig:GCN2Flow}}, we develop a prediction architecture, GCN2Flow that comprises several Temporal Convolutional blocks (TC blocks) and Spatial GCN blocks (SGC blocks). In this section, the stop network based on routes is defined as $\mathcal{G}_{stop}=\left\lbrace \mathcal{V}_s, \mathcal{E}_s, \mathcal{A}_s \right\rbrace$, where $\mathcal{V}_s$ is the set of stop nodes and $\mathcal{E}_s$ is the set of edges. $e^s_{ij}\in\mathcal{E}_s$ indicates the existence of a route from the stop $i$ to the next stop $j$. $\mathcal{A}_s$ is a weighted adjacency matrix, whose value of elements denote the geographical distance.

\par On the one hand, passenger flow prediction leverages historical time series data i.e., the past $t$ time steps are used to predict the next time step. Recurrent neural network-based methods are popular in time-series prediction. However, they have issues, such as time-consuming training and insensitive to the dynamics of long sequences. Therefore, in our TC block, we define a temporal gated convolutional operation, which utilizes gated linear units (GLU)~\cite{Dauphin2017Language} as a non-linearity with the residual connection. We assume that $C_{in}$, $C_{out}$ are respectively the number of input and output channels, and $X \in \mathbb{R}^{t \times |\mathcal{V}_s| \times C_{in}}$ is the input of the TC block. The operation is as follows:
\begin{equation}
\begin{split}
	&TC(X)=ReLU((XW_0+b_0)\otimes sigmoid(XW_1+b_1)\\
	&\quad+(XW_2+b_2)),
\end{split}
\end{equation}
where $W\in \mathbb{R}^{k\times C_{in}\times C_{out}}$ (k is the size of  convolutional kernel), $b\in\mathbb{R}^{|\mathcal{V}_s|\times C_{out}}$ are learned parameters, and $\otimes$ is the element-wise product between matrices.

\par Note, there are spatial connections between stops of bus routes. For example, the passenger flow of a stop is related to the flow of its neighboring stops. Therefore, the spatial graph convolutional operation through Chebyshev polynomials and first-order approximation~\cite{kipf2017semi} can be written as:
\begin{equation}
    SGC(X)=\phi\left( \tilde{D}^{-\frac{1}{2}}\tilde{A}_s\tilde{D}^{-\frac{1}{2}}XW \right),
\end{equation}
where $\tilde{A}_s=\mathcal{A}_s+I$, $\tilde{D}$ is degree matrix of the adjacent matrix $\mathcal{A}_s$, and $W$ is the learnable parameter matrix.

\par Therefore, one TC block and one SGC block are jointly utilized to extract Spatio-temporal features. The whole computational process of two blocks in l-th layer is designed as:
\begin{equation}
  F^{(l)}_j=TC(ReLU(SGC(TC(F^{(l-1)}_j)))),
\end{equation}
where $F^{(0)}_j=F_j\in\mathbb{R}^{t \times |\mathcal{V}_s| \times C^0_{in}}$ ($C^0_{in}=1$ in this case) is the flow matrix of stops with $t$ time steps in pattern $j$. $ReLU(\cdot)$ is the activation function (Rectified linear unit). Furthermore, we execute an extra temporal gated convolutional operation and attach a Fully-Connected (FC) layer as the output of the whole network. Therefore, $\hat{F}_j=FC(TC(F^{(L))})$ is the prediction flow matrix of next time step. Therefore, the loss function for passenger flow prediction in pattern $j$ can be defined as:
\begin{equation}
\mathcal{L}=\left\| \hat{F}_j-F_j\right\| ^2
\end{equation}

\par Finally, we train multiple GCN2Flow models, the number of which depends on the number of passenger mobility patterns. Therefore, each mobility pattern has its own GCN2Flow model to predict its passenger flow, and then
we merge prediction result of all GCN2Flow models to obtain the total passenger flow prediction result. Algorithm \ref{alg:MPGCN} presents the pseudocode for the training and predicting process of MPGCN.
\begin{algorithm}
	\caption{MPGCN predicts passenger flow}
	\label{alg:MPGCN}
	\begin{algorithmic}[1]
		\REQUIRE $m$: The number of passenger patterns;\\
		$F_j$: the matrix of stop passenger flow in the pattern $j$; \\
		$A_s$: the adjacent matrix of stop network; \\
		$T$: the time-steps of flow sequence;
		\ENSURE $\hat{F}$: the sequence of total flow prediction results;
		\STATE $normaliztion(A_s)$;
		\FOR{$j \gets 1$ to $m$}
			\STATE $normaliztion(F_j)$;
			\STATE $X, y \gets Samples(F_j, T)$; $\backslash\backslash$ obtain training set;
			\STATE building $GCN2Flow_j$ network;
			\STATE $GCN2Flow_j \gets GCN2Flow_j.train(X, y, A_s)$;
			\STATE $\hat{F}_j \gets unnormaliztion(GCN2Flow_j.predict)$;
			\STATE $\hat{F} += \hat{F}_j$;
		\ENDFOR
		\RETURN $\hat{F}$;
	\end{algorithmic}
\end{algorithm}

\subsection{Route Optimization} \label{route_optimization}
\par Finally, we use the prediction result of MPGCN to attempt a simple application case study. Once ensuring the accuracy of passenger flow prediction, we can assume that the prediction result is the real flow distribution of bus stops at the next time interval. In our framework, we show an example, route optimization, which is closely related with flow prediction, and our optimization task aims at providing a new travel route to avoid crowded bus stops, thus relieving the pressure of overcrowded bus stops in the public transportation systems. 

\par Therefore, we mainly focus on passenger diversion and ride experience as the optimization objective $O$, measured by the standard deviation ($std$) of traffic flow of all bus stops. More specifically, given the Origin-Destination ($OD$) matrix of passengers (shown in Section \ref{experiments}) and bus route network, how to select optimal routes from the candidate route set becomes the primary purpose. Mathematically, the objective function, $O(f)$, is similar with the standard deviation calculation formula, which means $\nabla^2O(f) \ge 0$ and the domain of variable $f$ is a finite set. Then, the optimal route selection problem can be abstracted into a convex optimization problem. In other words, given a stop network $\mathcal{G}_{stop}$ and traffic condition at the next time interval, the objective is to minimize the $std$ of traffic flow of all bus stops by changing the travel route of some passengers. Therefore, the objective function and constraints can be defined as follows:

\begin{equation}
\label{eq:objective}
	\min_{f_1, f_2, \ldots, f_{|\mathcal{A}_s|}} O(f)=\sqrt{\frac{1}{|\mathcal{A}_s|}\sum_{i=1}^{|\mathcal{A}_s|}(f_i-\bar{f})^2},
\end{equation}

\begin{equation}
\label{eq:constraint}
	\begin{split}
	s.t.,& \left\{
		\begin{aligned}
		&RS=g_1(\mathcal{G}_{stop}, OD, \hat{F})\\
		&F^*=g_2(r),\\
		&f_i\in F^*, \\
		&t\in T, \\
		&r\in RS,\\
		&len(r)-len(r_{shortest}) \le \epsilon
		\end{aligned}
	\right.
	\end{split}
\end{equation}
where function $g_1()$ can generate candidate routes finite set, $RS$, by passing in arguments including $\mathcal{G}_{stop}$, $OD$ matrix at time $t$ and the predicted passenger flow at the next time $t+1$, and $g_2()$ counts the passenger flow matrix of all stops based on a candidate route in $RS$. Besides, in the process of generating candidate routes set, we would set a threshold $\epsilon$ to ensure minimal additional cost of passenger travel time, that is, $len(r)-len(r_{shortest}) \le \epsilon$ ($\epsilon=5$ in this part), $r\in RS$, where $r_{shortest}$ is the shortest travel route based on $OD$. As a result, we can obtain the optimal routes of passengers, which can make the passenger flow of bus stops more balanced and relieve the crowded situation on the bus to a certain extent.

\subsection{Complexity Analysis}
\par In the part of deep clustering, we denote $d_l$  as the dimension of the input of $l-th$ layer in the autoencoder. The time complexity of the autoencoder is $\mathcal{O}(|\mathcal{V}_p||X|d_1^2\cdots d_L^2)$, and the time complexity of GCN module is $\mathcal{O}(|\mathcal{E}_p||X|d_1\cdots d_L)$. Besides, we assume that there are K clusters, and the time complexity of \eqref{eq_cluster} is $\mathcal{O}(|\mathcal{V}_p|K+|\mathcal{V}_p|log|\mathcal{V}_p|)$. Therefore, the total time complexity is the sum of the above three, and is linearly related to the $|\mathcal{V}_p|$ and $|\mathcal{E}_p|$. Similarly, the time complexity of flow prediction method is $\mathcal{O}(|\mathcal{E}_s||\mathcal{V}_s|d_1\cdots d_L)$.

\section{Experiments} \label{experiments}
\par To demonstrate the efficiency of our proposed MPGCN, we conducted a series of experiments. In this section, firstly, we describe the experimental dataset in detail, including preprocessing of data, analysis of data, and sharing-stop network. Secondly, we show the related parameters setting in experiments. Next, we present the analysis of mobility patterns and the effectiveness evaluation of the prediction performance compared with other methods. Finally, we suggest the application value of our prediction results through a case study.

\subsection{Data Description and Analysis}
\subsubsection{Data Description}
\par The real-world bus dataset is employed in our experiments, a typical kind of Spatio-temporal data, shown in \tablename{ \ref{tbl:record}} and \tablename{ \ref{tbl:arriving_leaving}} that contains bus record dataset from bus card, credit card, and Qr code, and bus stop arriving-leaving dataset comprising 12 used fields, which can basically cover the majority of bus passengers and reflect the overall trend of passenger mobility. Besides, the information about bus stops includes longitude, latitude, and the sequence of bus stops in the route. The bus dataset was generated by buses in Jiangsu, by Panda Bus Company, for 30 days (nine weekend days and 21 weekdays) from November 1st to 30th, 2019, and 18 hours a day from 05:00 to 23:00. It is noted that we do not use passengers' private information, so there is no privacy issue with our data.

\begin{table}
	\caption{Description of the bus record dataset}
	\label{tbl:record}
	\begin{tabular}{lll}
		\hline
		Field & Annotation & Examples\\
		\hline
		bus\_no & ID of each bus & 11180 \\
		card\_no& ID of each passenger & 2230000010282075 \\
		cardType& Payment card type & 1 \\
		riding\_time & Record time and date & 2019-11-01 05:29:20 \\
		routeId & ID of each route & 106\\
		\hline
	\end{tabular}
\end{table}

\begin{table}
	\caption{Bus stop arriving-leaving dataset details}
	\label{tbl:arriving_leaving}
	\begin{tabular}{lll}
		\hline
		Field & Annotation & Examples\\
		\hline
		bus\_no & ID of each bus & 61189 \\
		enterTime& Enter stop time and date & 2019-11-01 06:37:59 \\
		leaveTime& Leave stop time and date & 2019-11-01 06:38:12 \\
		stopId & ID of each stop & 46976 \\
		routeId & ID of each route & 157 \\
		directId & 0 for upline, 1 for downline & 0 \\
		stayTime & Bus waiting time (seconds) & 13 \\
		\hline
	\end{tabular}
\end{table}

\begin{figure}[htbp]
	\centering
	\includegraphics[width=0.45\textwidth]{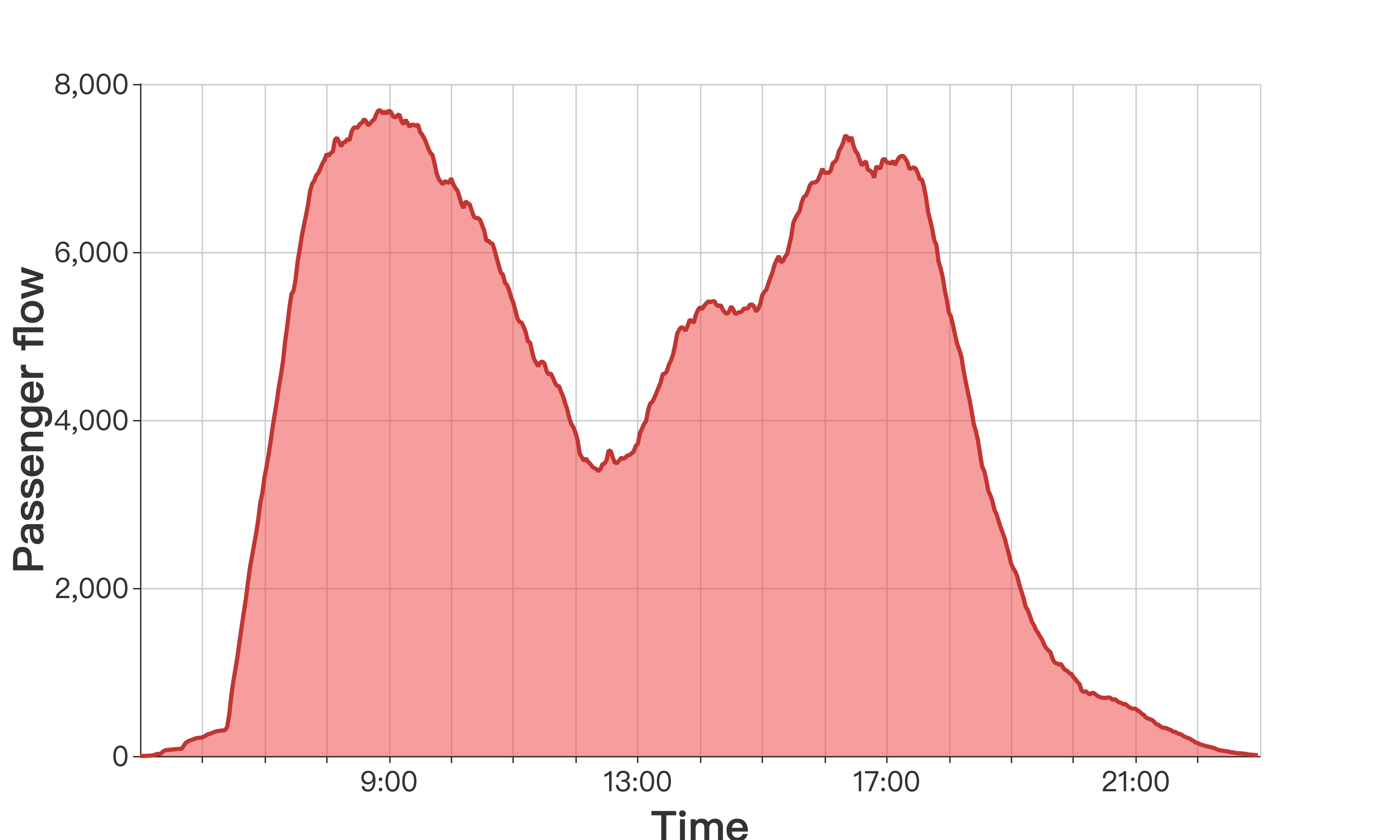}
	\caption{A example of passenger flow trend on November 1st, 2019.}
	\label{one_day_flow}
\end{figure}

\subsubsection{Preprocessing and Analysis}
\par From the raw data, we could not directly obtain bus stops where the passengers get on the bus and scan their card or Qr code. Hence, we first need to match bus stops. Considering the real-world experience of bus riding in China, the bus may have already run when passengers scan their card or Qr code. This fact implies the scanning time may not be between entering and leaving time and thus has a certain deviation. Consequently, we empirically selected 20 seconds to increase the time interval to expand the matching time range. Algorithm \ref{alg:matching} presents the technique for matching stops. Another difficulty with data preprocessing is that the drop-off stops of passengers are not given, unlike subway travel, which means the destination of passengers is not clear. Usually, travel by bus is symmetrical. For example, passengers' origin stop and destination will be switched to return to the origin stop. Based on this assumption, we will extract all getting-on stops and corresponding bus lines, and the two stops will be regarded as the origin and destination if these two stops are on the same line for a certain passenger. Therefore, the symmetrical $OD$ matrix can be inferred.
\begin{algorithm}
	\caption{Matching stops}
	\label{alg:matching}
	\begin{algorithmic}[1]
		\REQUIRE $Ride\_Records$: The table based on bus record data; \\
			$Bus\_Records$: the table based on bus stop arriving-leaving data; \\
			$\tau$: the time interval;
		\ENSURE $P2S$: the dictionary of matching stop;
		\STATE Load $Bus\_Records$; $\backslash\backslash$ convert to specific data structure for easy retrieve;
		\FOR{$i$ in $Ride\_Records$}
			\STATE	$temp \gets Bus\_Records.where(i)$ $\backslash\backslash$ obtain retrieve information based on bus id and date;
			\FOR{ $j$ in $temp$}
				\IF{$ i.riding\_time.time \ge j.enterTime-\tau$ and $ i.riding\_time.time \le j.leaveTime +\tau$}
					\STATE $\backslash\backslash$ matching condition;
					\STATE add $j.stopId$ to $P2S[i.card\_no]$;
					\STATE break;
				\ENDIF
			\ENDFOR
		\ENDFOR
		\RETURN $P2S$;
	\end{algorithmic}
\end{algorithm}

\begin{figure}[htbp]
	\centering
	\subfigure[]{\includegraphics[width=0.22\textwidth]{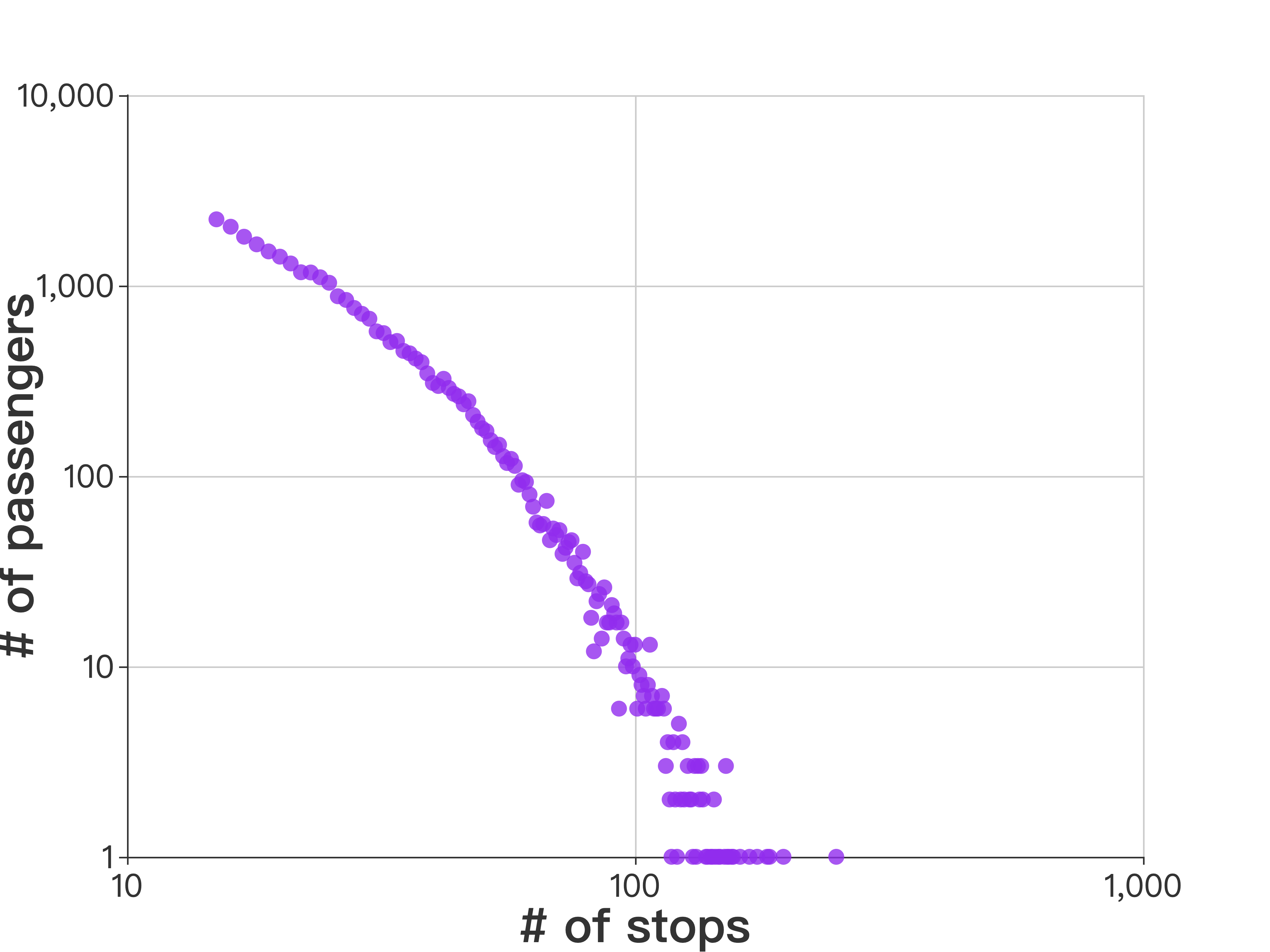}}
	\subfigure[]{\includegraphics[width=0.22\textwidth]{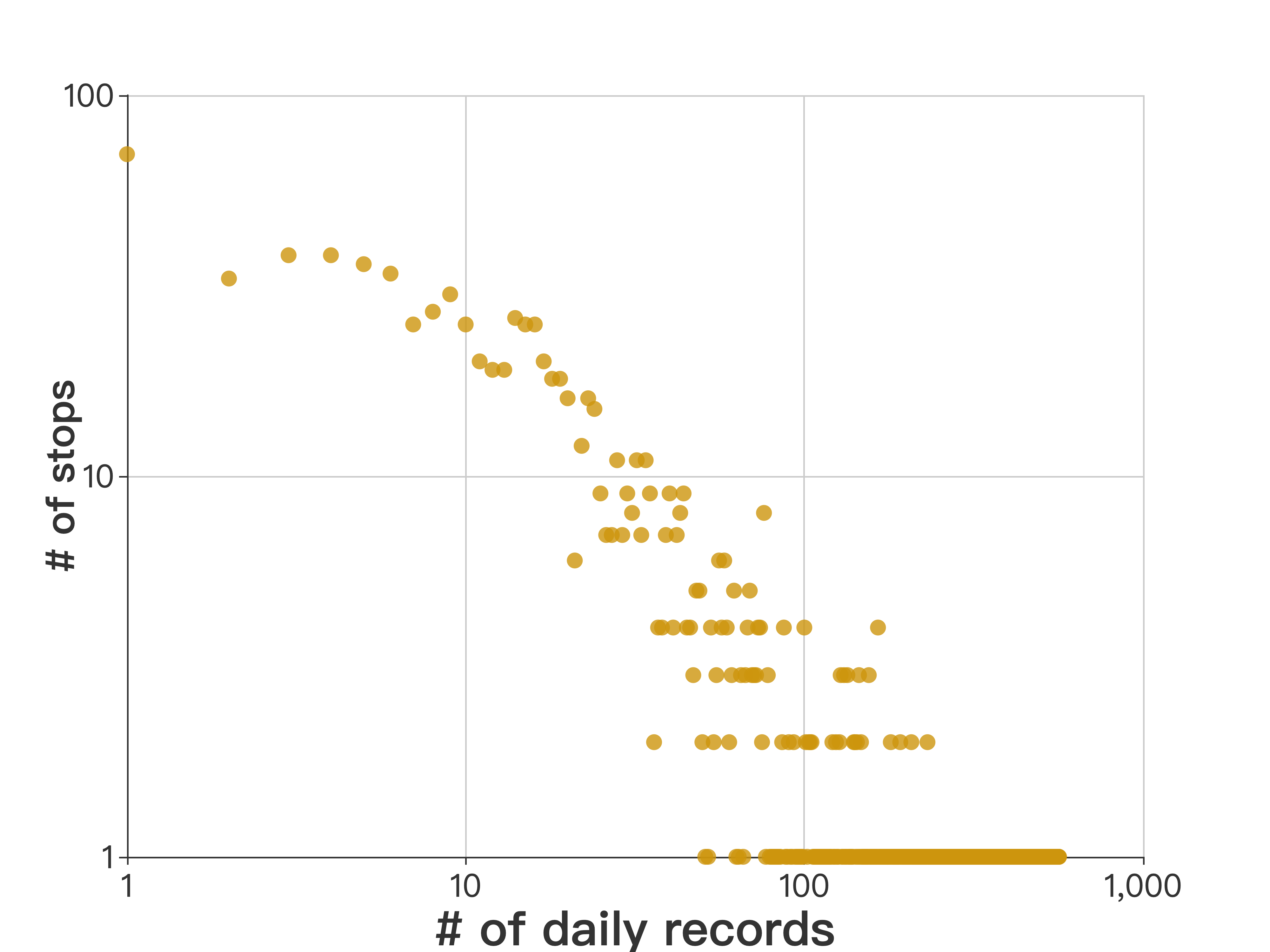}}
	\caption{The heavy-tailed distribution of the number of passengers, who pass a certain number of stops (a), and the number of stops that have a certain number of records (b).}
	\label{fig:p_s_distribution}
\end{figure}

\begin{figure}[htbp]
	\centering
	\includegraphics[width=0.45\textwidth]{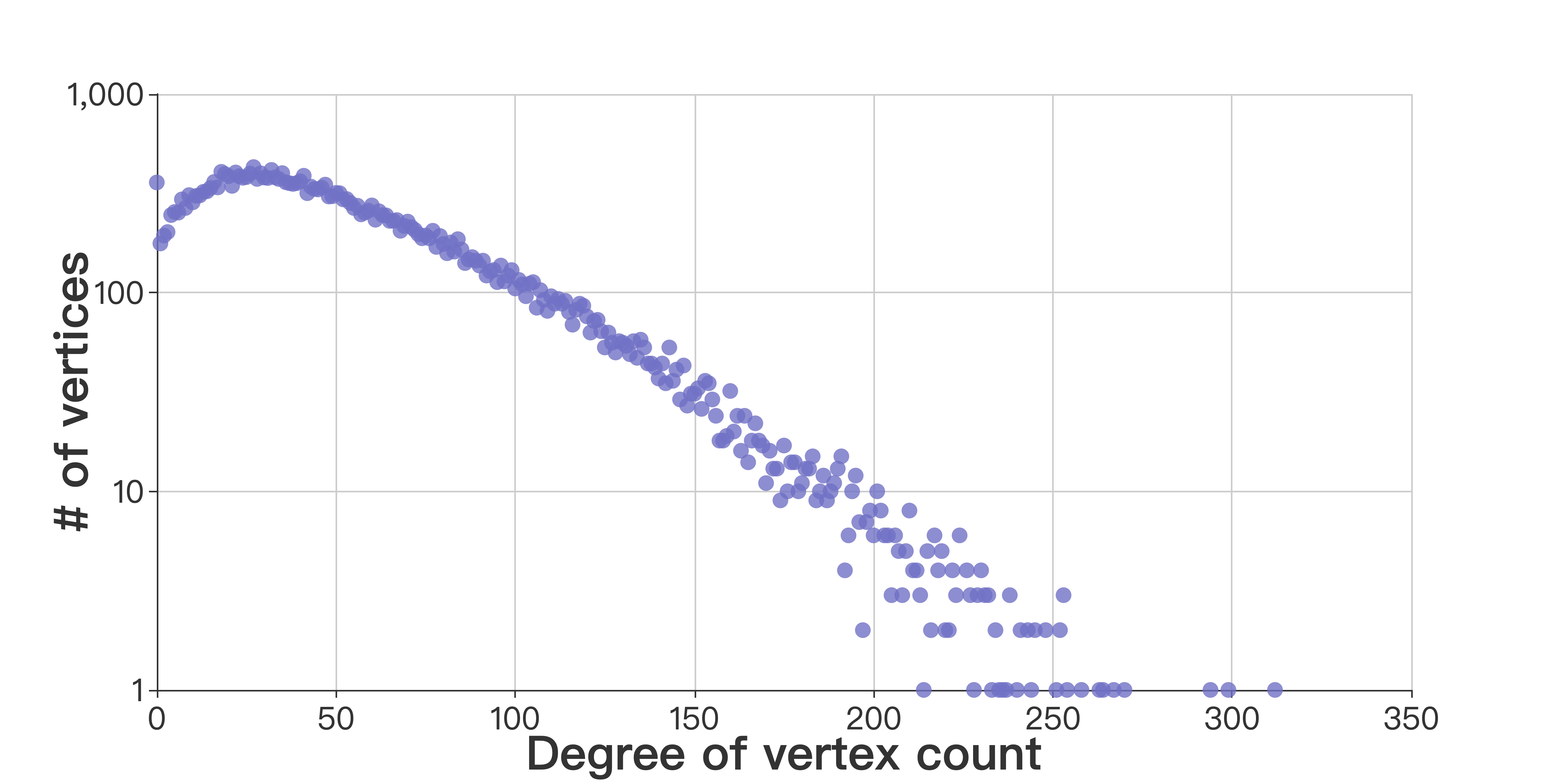}
	\caption{The distribution of degrees of all passenger nodes.}
	\label{fig:G_p_degrees}
\end{figure}

\par After matching stops and inferring the $OD$ matrix, we expand the $OD$ matrix into stop trajectories, which enable us to count passenger flow at each stop. Besides, the number of records per passenger is required to be higher than a particular value. This higher number of records is required because of our demand for exploring passenger mobility and its laws. For this, some passengers and their records are needed to be filtered. For example, passengers having a few records in a month or starting to generate a card record end of the month were filtered. Also, we found data anomalies like the missing value problem among the raw data. For example, bus stop arriving-leaving data of some buses for several days is missing, influencing model training. Hence, inspired by~\cite{mcleod2007estimating}, we use the linear interpolation method to reduce errors during the computing of traffic flow. The matching process screens some records having less impact on passenger mobility laws. Finally, the data extracted by us contains 857900 bus records of 31353 passengers, 214 routes, and 1114 stops in Huai'an city.

\par Apart from data preprocessing, cleaning, and filtering, we analyze the relevance of passenger flow and time. In \figurename{ \ref{one_day_flow}}, we list passenger flow for one day (November 1st, 2019). \figurename{ \ref{one_day_flow}} shows daily flow is similar in shape i.e., there are 2 or 3 peaks and 1 or 2 troughs. Therefore, this similarity also means that not only is traffic flow regular, but also that passenger mobility follows the law. Besides, as shown in \figurename{ \ref{fig:p_s_distribution}}, we calculate the number of stops that each passenger passes and the average daily number of records at each stop in a month and describe their distribution based on a certain order of magnitude after counting them. \figurename{ \ref{fig:p_s_distribution}} exhibits both distributions conform with the heavy-tailed distribution, which is why our network embedding part is inspired. The existence of heavy-tailed distribution is the main reason for the latent effectiveness in the process of natural language model and network representation.

\par After preprocessing, the sharing-stop network is constructed with Algorithm \ref{alg:Construction}. As shown in \figurename{ \ref{fig:G_p_degrees}}, we count the degrees of all passenger nodes in the network and draw their distribution, which suggests the analogous distribution, an expected heavy-tailed distribution, and shows the potential law of passenger mobility.

\begin{figure*}[htbp]
	\centering
	\subfigure[time\_step=5 min]{\includegraphics[width=0.3\textwidth]{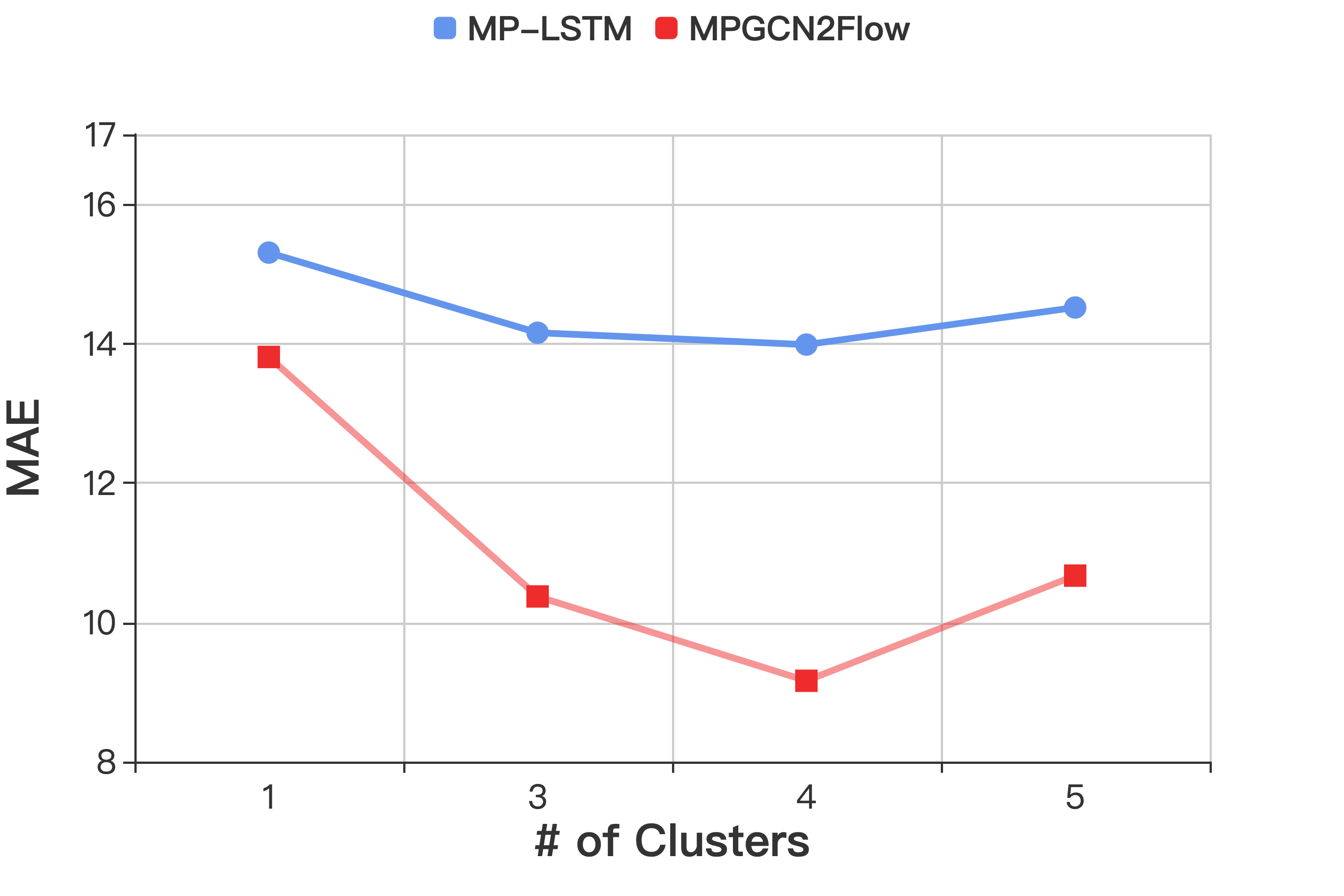}}
	\subfigure[time\_step=15 min]{\includegraphics[width=0.3\textwidth]{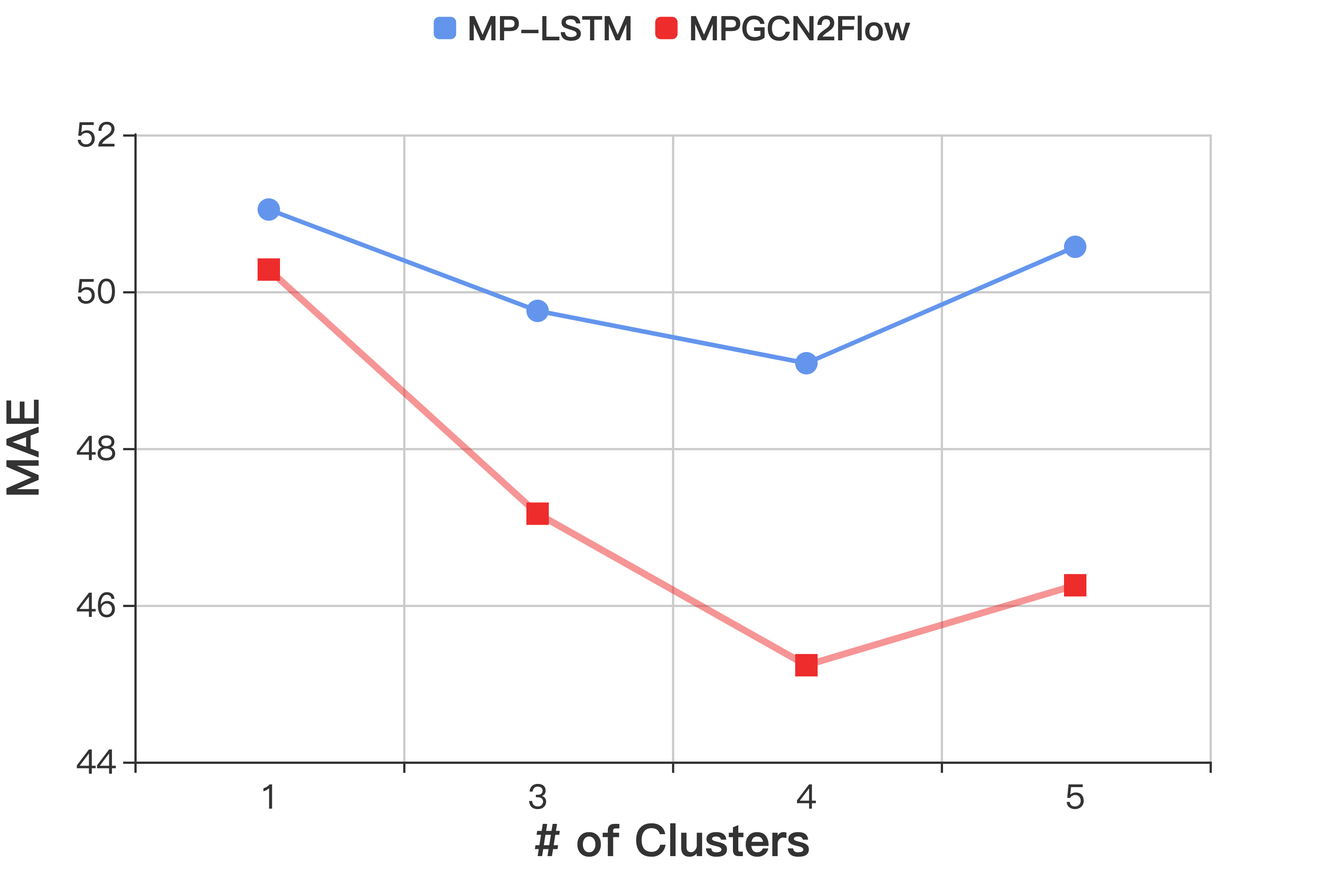}}
	\subfigure[time\_step=30 min]{\includegraphics[width=0.3\textwidth]{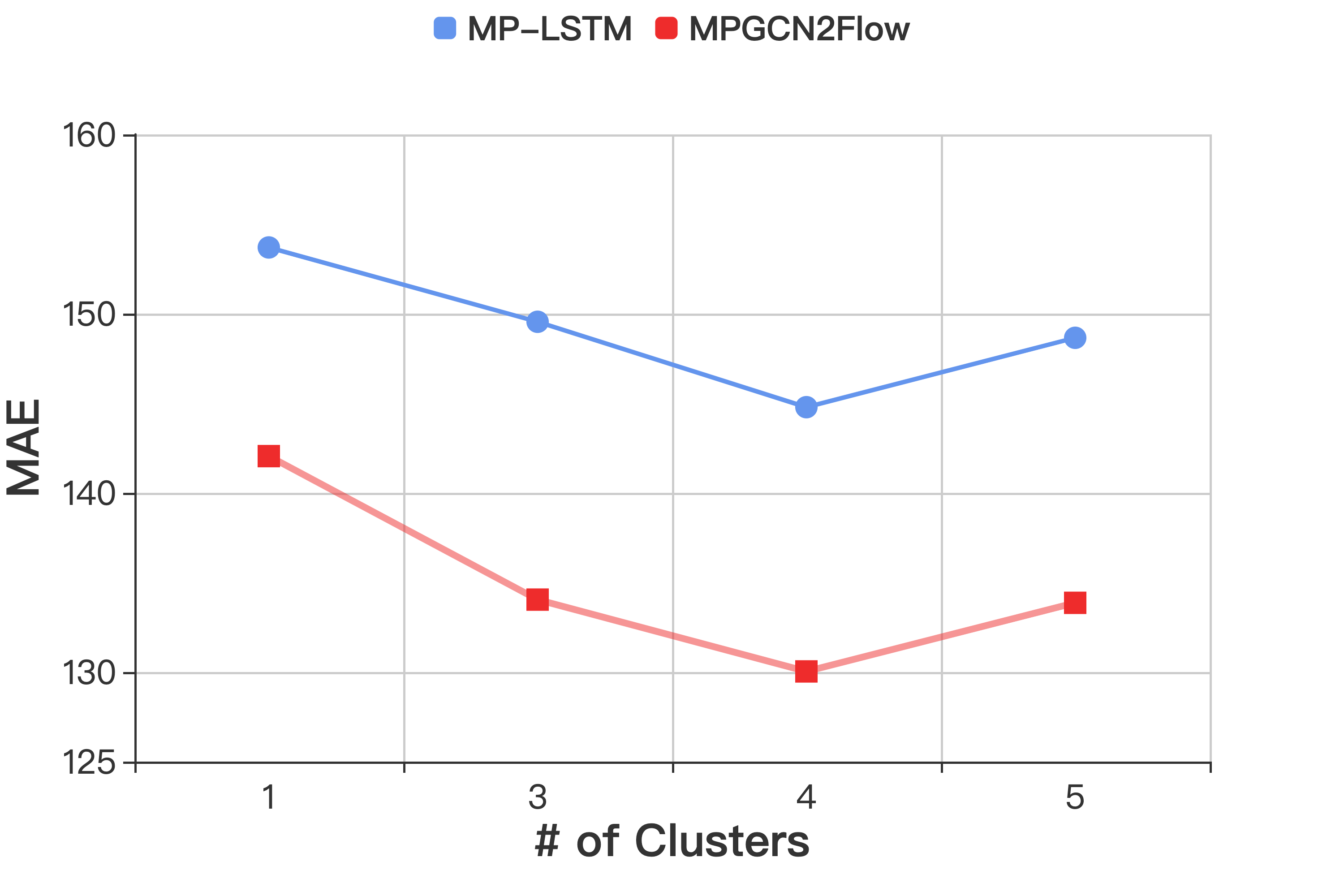}}
	\caption{The prediction results with different number of clusters, where $n\_cluster=1$ denotes prediction without passenger patterns. (a) time\_step=5 min. (b) time\_step=15 min. (c) time\_step=30 min.}
	\label{fig:analysis_n_c_flow}
\end{figure*}

\subsection{Experimental Settings}
\par In the part of deep clustering with GCN, we extracted passenger mobility patterns based on the sharing-stop network.  Since the size of the sharing-stop network is large, we set the dimension of the neural network of AE to $\left|\mathcal{V}_p\right|$-100-100-500-16, which was the same as the GCN module. In addition, the result of our method is insensitive to hyper-parameters, therefore the setting of hyper-parameters is $(\theta_1, \theta_2, \theta_3)=(1, 0.5, 0.05)$. The learning rate and epoch number were 0.001 and 100, respectively. For considering the impact of cluster number on the passenger flow prediction comprehensively, we set 3, 4, and 5 as the number of clusters (passenger mobility patterns), and selected the most suitable number of passenger mobility patterns to analyze their latent laws based on the prediction results.

For flow prediction, the historical passenger flow data for an hour were used as the input of our proposed method to predict the flow for the next time step, and the time step was taken as 5, 15, and 30 minutes, respectively. There were 5 TC blocks and 2 SGC blocks in our GCN2Flow. The convolution kernel size of both blocks was 3. The batch size and the learning rate were 64 and 0.001 with epochs 100, respectively. Before training, the flow of all bus stops was normalized with the Z-score method, and the stop network as Laplace matrix was also normalized.

Further, similar with previous works~\cite{yu2018spatio, Kong2018subway}, we utilized the following three most common metrics used in the comparison of passenger flow prediction methods: Mean Absolute Error (MAE), Root Mean Square Error (RMSE), and Correlation Coefficient (CC).

\begin{itemize}
	\item[-] Mean Absolute Error (MAE).
	MAE is the mean of all absolute errors between the predicted values and their corresponding real values, whose equation is given as follows:
	\begin{equation}
	MAE=\dfrac{1}{n}\sum_{i=1}^{n}\left|\hat{y_i}-y_i\right|
	\end{equation}
	where $\hat{y_i}$ and $y_i$ denote the predicted values and real values, respectively.
	\item[-] Root Mean Square Error (RMSE).
	RMSE measures the deviation between predicted values and their respective real values. RMSE is defined as:
	\begin{equation}
	RMSE=\sqrt{\dfrac{\sum_{i=1}^{n}\left|\hat{y_i}-y_i\right|^2}{n}}
	\end{equation}
	\item[-] Correlation Coefficient (CC).
	CC is used to verify the correlation between variables and has different forms. In this paper, we used Pearson CC to measure correlation possessing the following formula:
	\begin{equation}
	CC=\dfrac{cov(\hat{y_i}, y_i)}{\sqrt{var(\hat{y_i})\cdot var(y_i)}}
	\end{equation}
	where $cov(\cdot,\cdot)$ and $var(\cdot)$ represent the covariance and variance, respectively.
\end{itemize}

\par In addition, our model is implemented based on Pytorch framework, and the experiments is executed with NVIDIA RTX 2080 Ti.
\subsection{Analysis of Passenger Mobility Patterns}
\par To investigate the impact of extracting passenger mobility patterns on prediction, we varied the number of clusters used in prediction models. In \figurename{\ref{fig:analysis_n_c_flow}}, $n\_cluster=1$ indicates the GCN2Flow and LSTM~\cite{Sutskever2014lstm} models were directly used to predict passenger flow, while $n\_cluster$=3, 4, and 5 means that the MPGCN and Multi-pattern LSTM (MP-LSTM) models were used to predict flow of passengers for 3, 4, and 5 different mobility patterns, respectively. \figurename{ \ref{fig:analysis_n_c_flow}} shows it is effective to enhance the performance of prediction combining passenger mobility patterns i.e., MPGCN and Multi-pattern LSTM (MP-LSTM) are better than GCN2Flow and LSTM, respectively.

\par As a result, the MPGCN and Multi-pattern LSTM (MP-LSTM) models produced the best performance for $n\_cluster=4$. Therefore, we selected $n\_cluster=4$ as the number of passenger mobility patterns for further analysis. In our studies, a clustering result is viewed as a mobility pattern. Due to the basics of the sharing-stop network, we suspect that passenger nodes in the same pattern tend to have similar travel habits, like taking fixed and frequent bus stops or routes. In this way, the number of passengers in the four patterns is 11857, 10537, 3475, and 5484, which add up to 31353. 

\par
\begin{figure*}[htbp]
	\centering
	\subfigure[Pattern 1]{\includegraphics[width=0.24\textwidth]{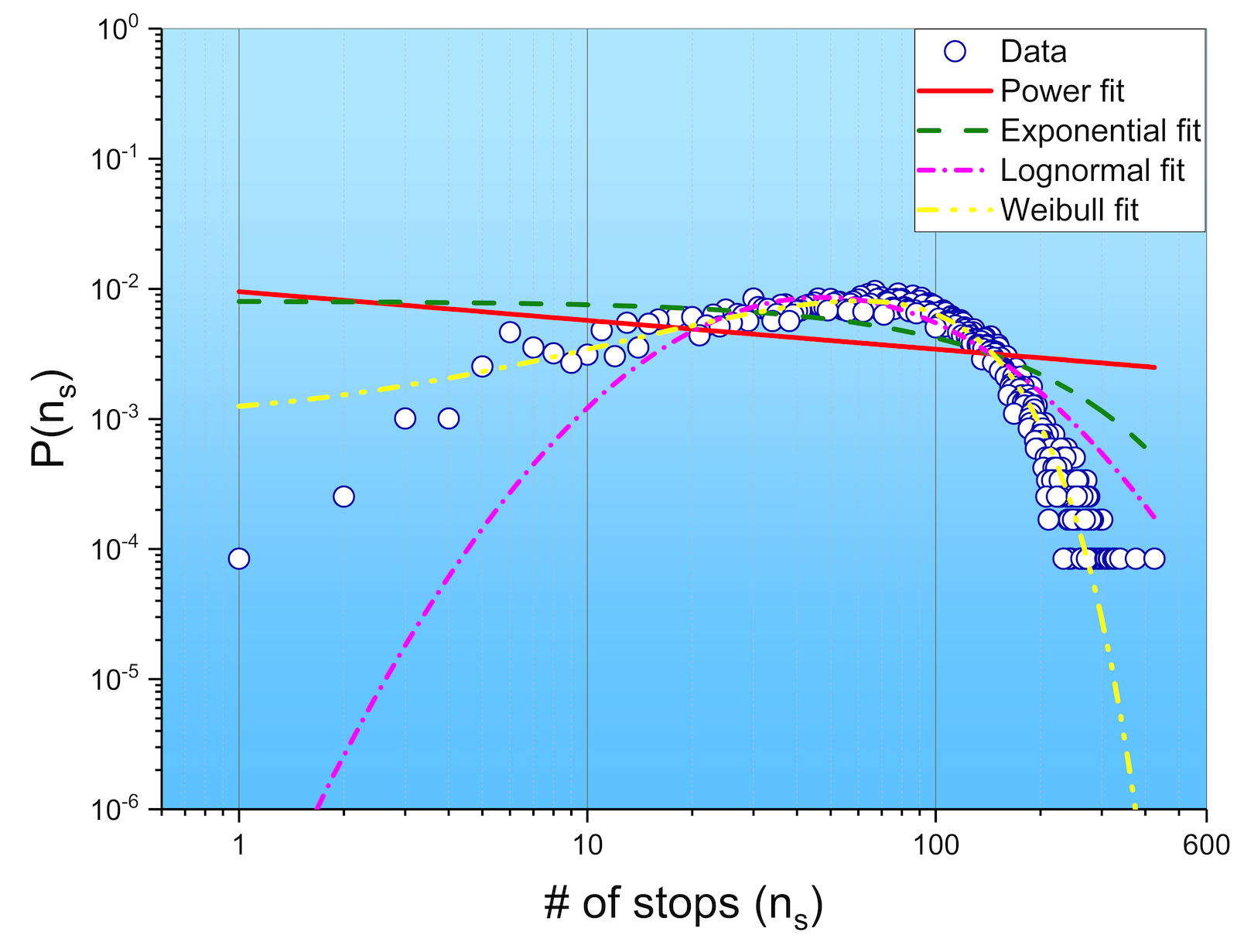}}
	\subfigure[Pattern 2]{\includegraphics[width=0.24\textwidth]{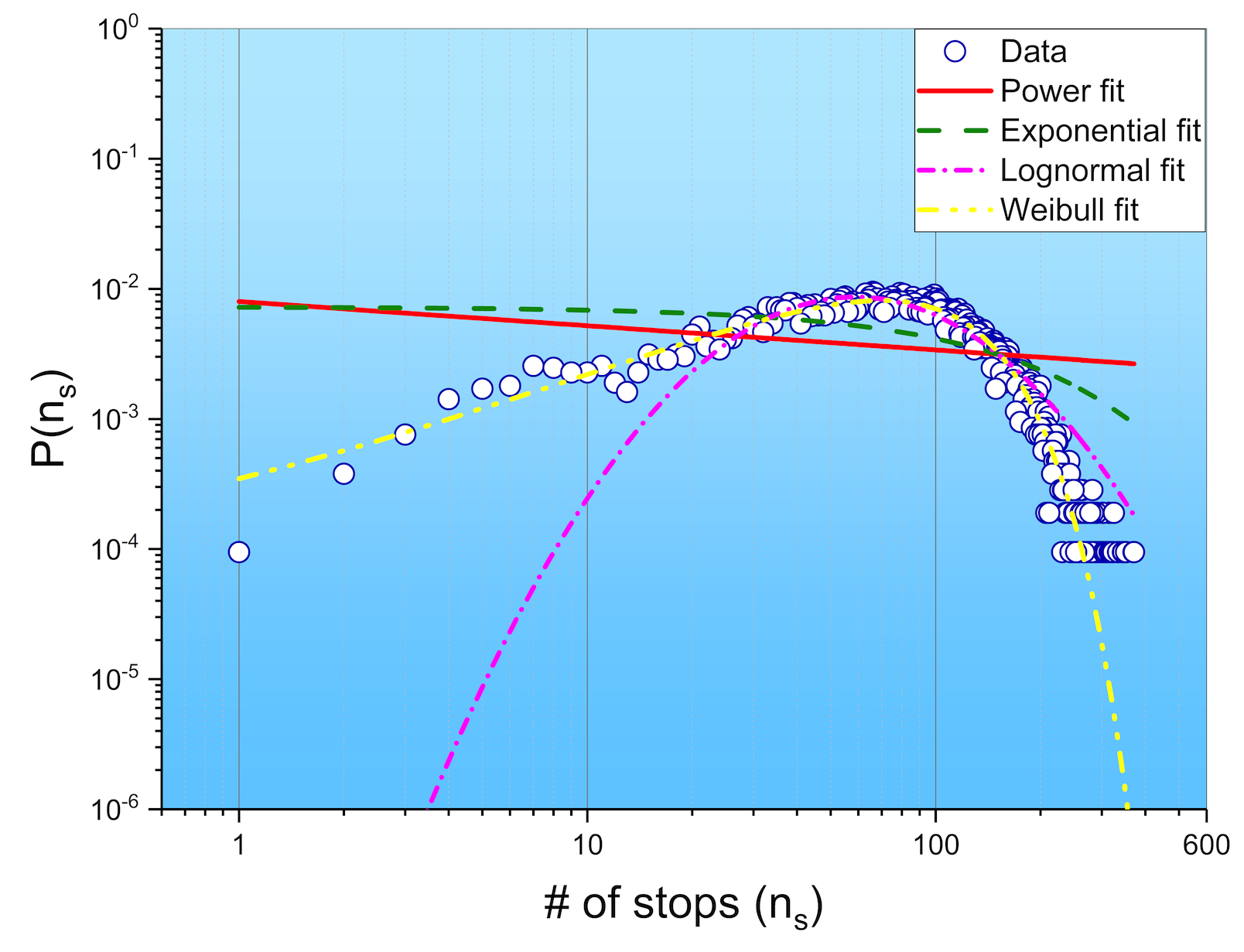}}
	\subfigure[Pattern 3]{\includegraphics[width=0.24\textwidth]{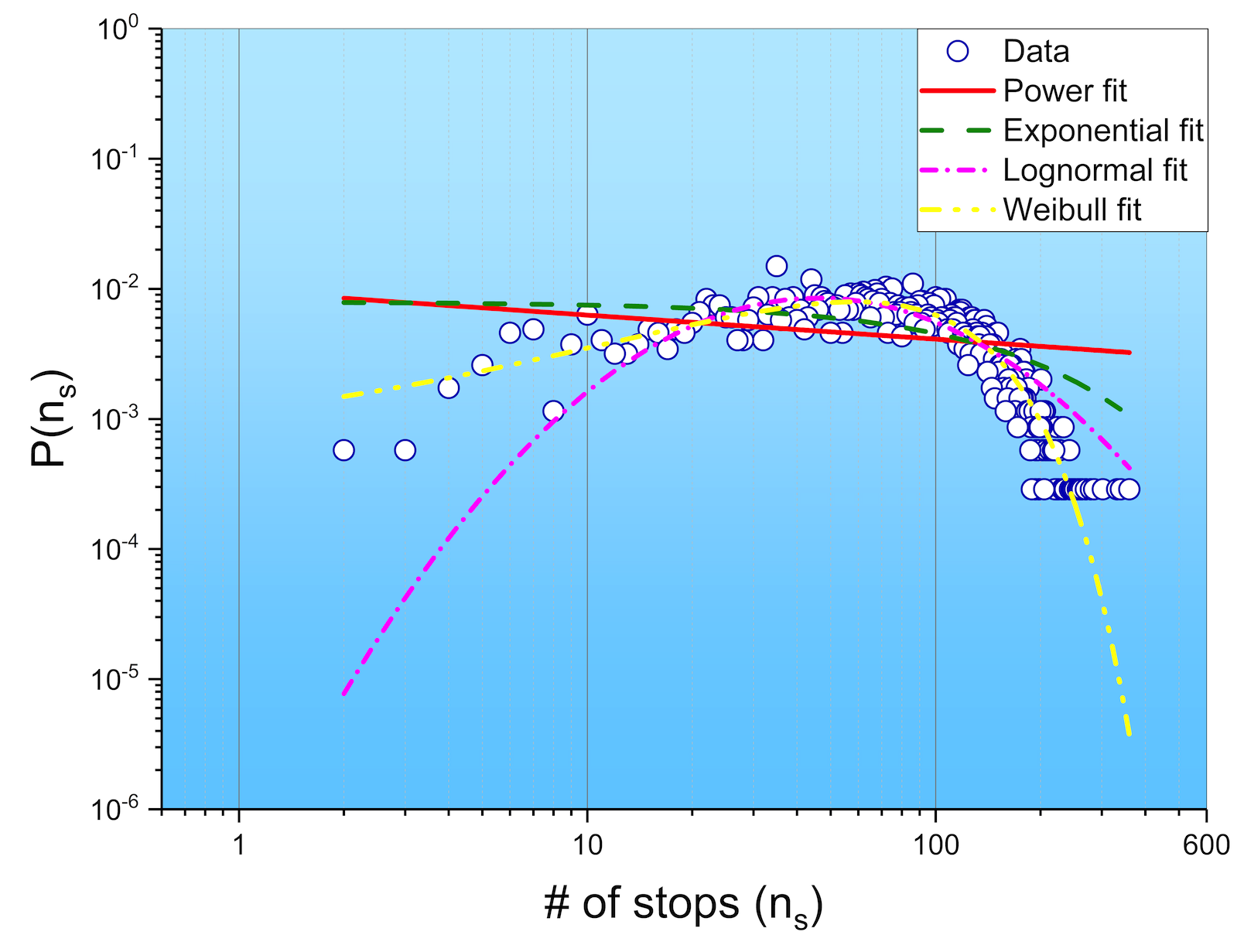}}
	\subfigure[Pattern 4]{\includegraphics[width=0.24\textwidth]{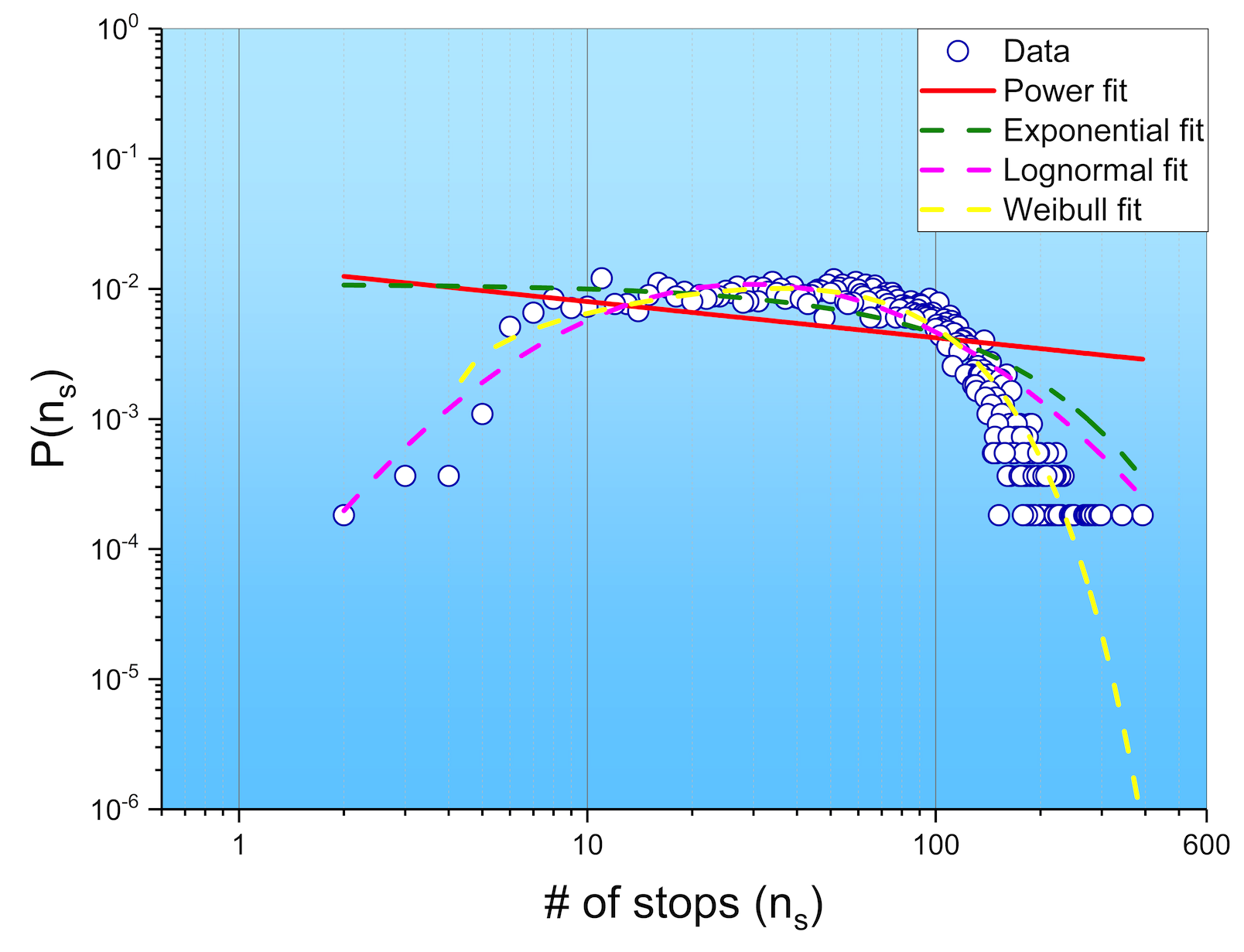}}
	\caption{Description about four mobility patterns of passengers with data fitness.}
	\label{fig:mobility_patterns_fitting}
\end{figure*}

 To further mine special laws hidden in mobility patterns, we show several heavy-tail distributions to fit the number of stops passengers pass ($n_s$), which are power-law, exponential, log-normal, and Weibull distributions. The probability density function (pdf) of these distribution are shown in \tablename{ \ref{tbl:distribution_pdf}}. From \figurename{ \ref{fig:mobility_patterns_fitting}}, for Patterns 1, 2 and 3, the distributions of $n_s$ have similar law i.e., log-normal and Weibull distributions are better fitting curves than the remaining two distributions (power-law and exponential). Before $P(n_s)$ reaches the highest value, the Weibull distribution has a better fitting effect, and after that, the log-normal distribution becomes better. Besides, by comparing key parameters of distributions, $c\simeq87, 91, 92$ in pdf of log-normal distribution; $a\simeq99, 102, 99$ and $r\simeq1.8, 1.9, 1.7$ in pdf of Weibull distribution, the similarity between log-normal and Weibull distributions can also be further confirmed. For Pattern 4 and $c\simeq71$ in pdf of log-normal distribution and $a\simeq73, r\simeq1.4$ in pdf of Weibull distribution, log-normal distribution achieves the best fitting. More specifically, through quantitative analysis, we note that 80 percent of passengers of Patterns 1, 2, and 3 tend to pass around less than or equal to 127 stops, while 80 percent of passengers of Pattern 4 only pass less than or equal to 101 stops. 

\par Furthermore, in order to verify our conjecture that passengers in the same pattern tend to have similar travel habits, we count the flow contribution of passengers of four mobility patterns in each bus route and analyze the distribution proportion of the top 40 bus routes in terms of total passenger flow. Then, we define the route preference of the pattern, which denotes the proportion of passengers of this pattern in the route exceeds 50\%. As shown in \tablename{ \ref{tbl:patterns_route_preference}} (see Appendix for details), noticeable preference differences of mobility patterns can be found. For example, passengers of pattern 4 contribute more than 90\% on routes 62, 63, 65, and 66. Particularly, for the routes not shown in \tablename{ \ref{tbl:patterns_route_preference}}, we also found that the proportion of pattern 1 and pattern 2 is relatively high and close, both of which are above 30\% or 40\%, such as routes 2, 4, 16, 26, and 31. This fact indicates that there is a specific shared similarity between the two mobility patterns in terms of travel. Therefore, through deep clustering, the effectiveness of our implicit mobility pattern extraction is verified by these analysis results. In other words, passengers of the same pattern tend to travel by bus on some fixed routes and contribute most of the traffic flow in those routes, while passengers with different patterns often choose different routes to travel. Besides, based on the latent laws of mobility patterns, it is effective to combine them with traffic flow prediction task.

\begin{table}
	\caption{Route preference of four mobility patterns}
	\label{tbl:patterns_route_preference}
	\centering
	\begin{tabular}{|c|l|}
		 \hline
		Pattern No. & \multicolumn{1}{c|}{List of Route Preference} \\ \hline
		1           & [5, 7, 12, 15, 18, 19, 32, 33, 36, 39, 53, 88, 89, 100]   \\
		2           & [1, 6, 11, 14, 22, 23, 28, 38, 116, K1]   \\
		3           & [50, 91, 713]   \\
		4           & [10, 20, 62, 63, 65, 66, 69] \\ \hline                            
	\end{tabular}
\end{table}

\begin{table}
	\caption{Heavy-tailed distributions}
	\label{tbl:distribution_pdf}
	\centering
	\begin{tabular}{|l|c|}
		\hline
		\multicolumn{1}{|c|}{Distribution} & Probability Density Function (pdf) \\
		\hline
		Power-law & $ax^b$\\[3pt]
		Exponential & $a\cdot \exp(bx)$ \\[3pt]
		Lognormal & $ \frac{A}{xw\sqrt{2\pi}}\exp[-\frac{(\ln (x/c))^2}{2w^2}]$ \\[4pt]
		Weibull & $ \frac{r}{a}(\frac{x-u}{a})^{r-1}\exp[-(\frac{x-u}{a})^{r}]$\\[3pt]
		\hline
	\end{tabular}
\end{table}

\begin{table*}
	\caption{Performance comparison of different approaches}
	\label{tbl:prediction_comparison}
	\centering
	\begin{tabular}{|c|c|c|c|c|c|c|c|c|c|c|c|}
		\hline
		\multicolumn{2}{|c|}{Methods}            & LR     & SVR    & GBDT   & ARIMA  & LSTM   & DCRNN        & STGCN       & \textbf{MP-LSTM} & \textbf{GCN2Flow} & \textbf{MPGCN}  \\ \hline
		\multirow{3}{*}{t\_step=5 min}  & MAE     & 21.56  & 22.62  & 22.85  & 23.90  & 15.30  & 13.93        & 14.01       & 13.98            & \underline{13.81}       & \textbf{9.16}   \\ \cline{2-12} 
		& RMSE(\%) & 35.14  & 37.45  & 37.85  & 43.68  & 23.76  & 21.28        & 22.25       & \underline{17.88}      & 20.41             & \textbf{15.82}  \\ \cline{2-12} 
		& CC      & 0.823  & 0.942  & 0.968  & 0.955  & 0.975  & 0.983        & 0.978       & 0.980            & \underline{0.986}      & \textbf{0.992}  \\ \hline
		\multirow{3}{*}{t\_step=15 min} & MAE     & 70.27  & 67.52  & 68.36  & 70.72  & 51.04  & 50.06        &  \underline{48.96} & 49.08            & 50.28             & \textbf{45.23}  \\ \cline{2-12} 
		& RMSE(\%) & 115.26 & 111.74 & 113.28 & 130.41 & 99.82  & 92.33        & 91.45       & 92.56            &  \underline{89.39}       & \textbf{81.33}  \\ \cline{2-12} 
		& CC      & 0.901  & 0.945  & 0.972  & 0.953  & 0.973  & 0.976        & \underline{0.979} & 0.976            & 0.978             & \textbf{0.984}  \\ \hline
		\multirow{3}{*}{t\_step=30 min} & MAE     & 164.69 & 155.73 & 155.92 & 161.99 & 153.69 & \underline{137.31} & 144.52      & 144.78           & 142.05            & \textbf{130.03} \\ \cline{2-12} 
		& RMSE(\%) & 283.58 & 270.42 & 294.71 & 321.23 & 279.82 & \underline{255.09} & 267.62      & 268.09           & 260.88            & \textbf{240.48} \\ \cline{2-12} 
		& CC      & 0.885  & 0.937  & 0.960  & 0.952  & 0.954  & \underline{0.968} & 0.964       & 0.964            & 0.966             & \textbf{0.975}  \\ \hline
	\end{tabular}
\end{table*}

\subsection{Passenger Flow Prediction}
\subsubsection{Prediction Result}
\par
\begin{figure}[htbp]
	\centering
	\subfigure[Weekday (Nov. 29th, 2019)]{\includegraphics[width=0.4\textwidth]{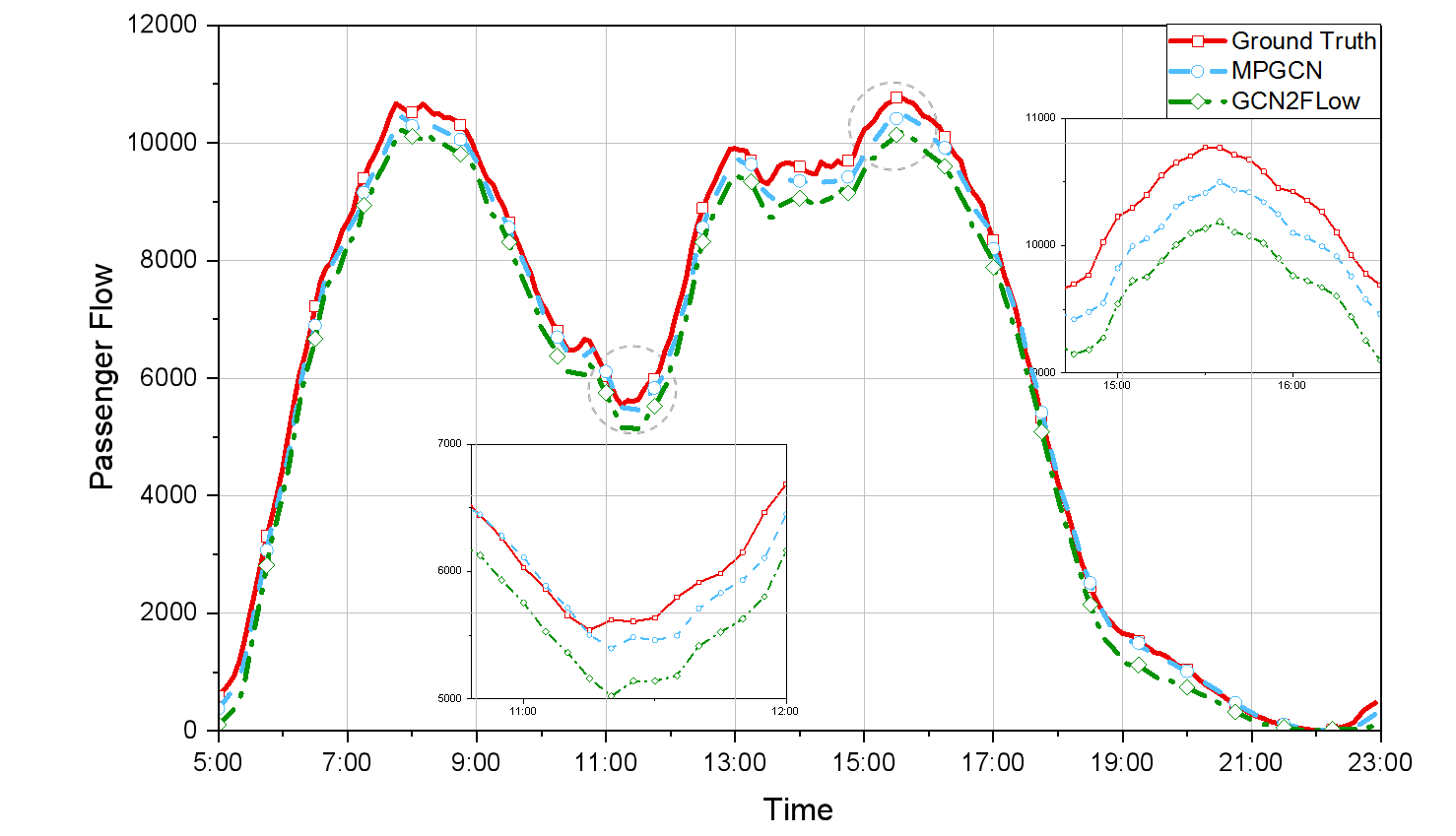}}
	\subfigure[Weekend day (Nov. 30th, 2019)]{\includegraphics[width=0.4\textwidth]{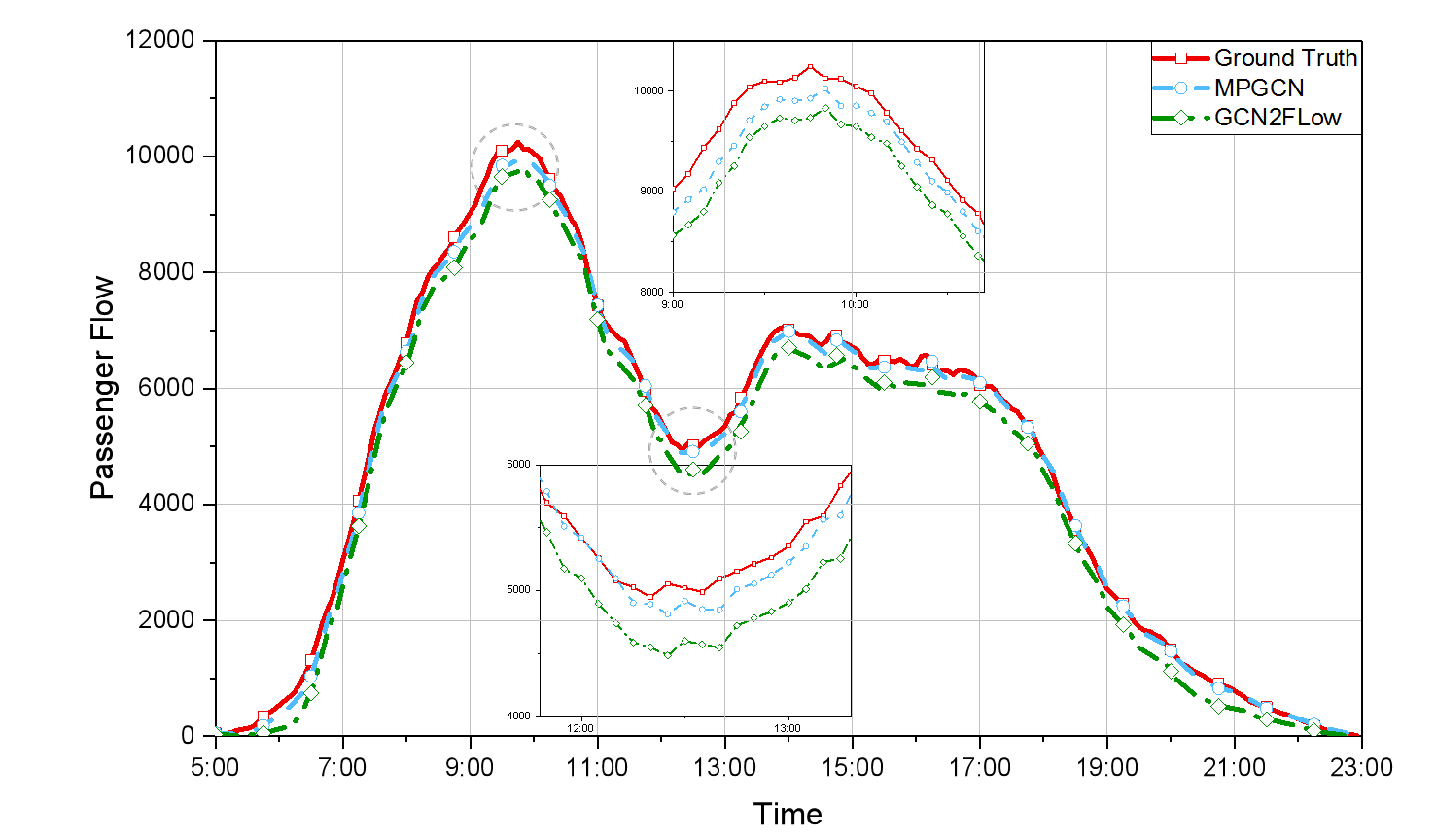}}
	\caption{Prediction results of MPGCN and GCN2Flow with time\_step=5 min including a weekend day (a), and a weekday (b). (Zoom in suitable view during a peak and a trough period)}
	\label{fig:prediction_results}
\end{figure}

 Based on the passenger mobility patterns obtained from the previous experiments, using GCN2Flow (MPGCN model described in Algorithm \ref{alg:MPGCN} for individual patterns), we predicted the passenger flow for each pattern. \figurename{ \ref{fig:prediction_results}} shows the short-term passenger flow prediction results of our proposal GCN2Flow and MPGCN with $time\_step=5$ for a weekday and a weekend day. The comparison between the predicted flow with the real flow justifies that the excellent prediction result is achieved. For different trends of weekdays and weekend days, our models both capture the features of passenger flow trends i.e., flow peaks and troughs. In terms of spatial features, the SGC block is capable of fast predicting the dynamic flow changes in the stop network based on the bus route network. Moreover, by combining passenger flow predictions using MPGCN, the prediction accuracy can be improved.

\subsubsection{Comparison of prediction approaches}
\par To verify the potential capability of our approach, we predicted passenger flow using the same dataset by many contemporary and popular relevant models that include machine learning models, mathematical-statistical model, and neural networks model: Logistic Regression (LR), Support Vactor Regression (SVR), Gradient Boosting Decision Tree (GBDT), ARIMA, LSTM~\cite{Sutskever2014lstm}, DCRNN~\cite{li2018dcrnn}, STGCN~\cite{yu2018spatio}, MP-LSTM (LSTM with passenger mobility patterns).

\par The results of prediction evaluation are presented in \tablename{ \ref{tbl:prediction_comparison}}. Our proposed MPGCN achieves the best performance in all three evaluation metrics. In general comparison, because of the complexity of data, the mathematical-statistical model, ARIMA, is the worst at predicting. Machine learning models have similar prediction performance, i.e., they can predict well for short-term flow, while their prediction accuracy is seriously reduced for the long-term without considering the relationship of spatial geographic information. Although the prediction performance of the neural networks model, LSTM, is respectful, which has better prediction results than those for the mathematical-statistical model and other machine learning models, it is still inferior to our models. It's worth mentioning that the results of STGCN and DCRNN are close to our GCN2Flow. Besides, in terms of passenger mobility patterns, when applying them in the prediction model (MP-LSTM), its performance can be further enhanced. In particular, for the $time\_step =30$ setting, MPGCN achieves \textbf{5.3\%} ($137.31\rightarrow130.03$) MSE reduction compared to other baselines. To sum up, the ability of GCN2Flow and the effectiveness of combining passenger mobility patterns, MPGCN vindicate their application in the passenger flow prediction.

\begin{figure}[htbp]
	\centering
	\subfigure[]{\includegraphics[width=0.4\textwidth]{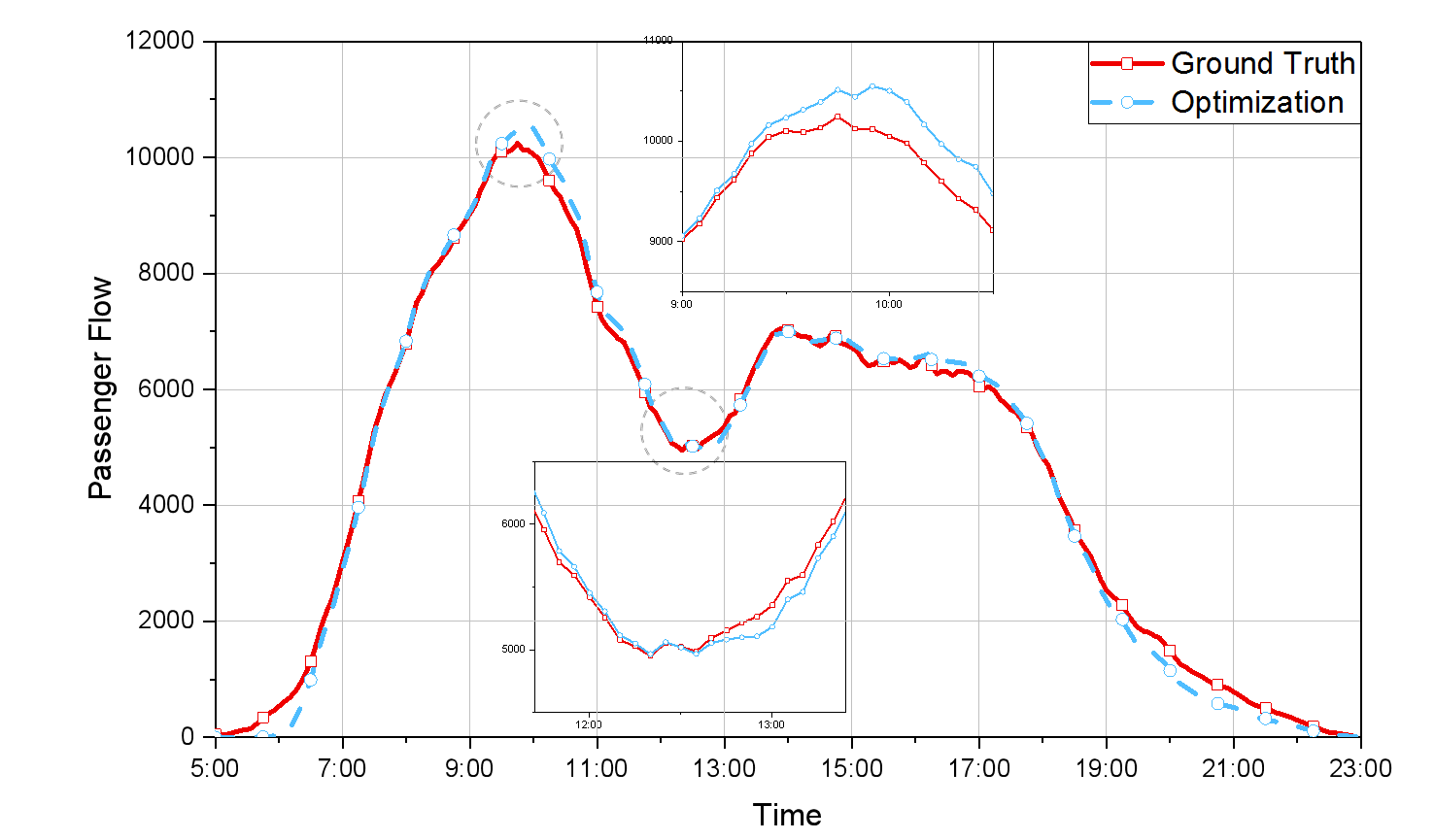}}
	\subfigure[]{\includegraphics[width=0.4\textwidth]{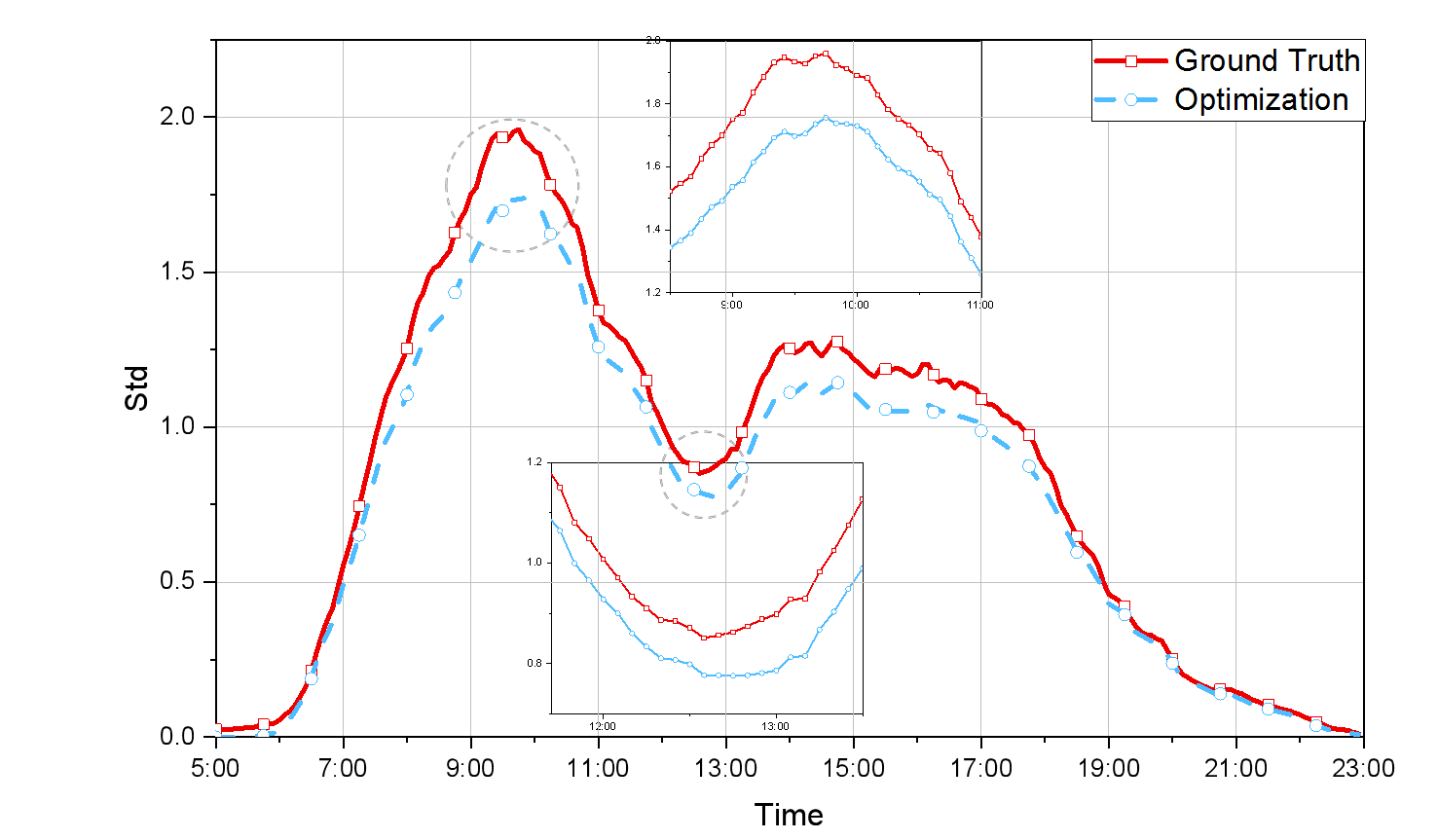}}
	\caption{The route optimization results compared with : the trend of (a) passenger flow, and (b) standard deviation of normalized flow. (Zoom in suitable view during a peak and a trough period)}
	\label{fig:optimization_route}
\end{figure}

\subsection{A Case Study of Route Optimization}
\par On the one hand, passenger flow prediction can be supplied to many downstream tasks. On the other hand, the feedback results of downstream tasks can also verify the accuracy of prediction. In our case, we define a route optimization task for allocating passenger flow and improving travel experience on the bus, which demonstrates the value of our proposed MPGCN in another way.

\par In the experiment, based on the OD matrix, we utilize prediction results for recommending an optimal route to each passenger, which meets objective function and constraint conditions (Eq. \ref{eq:objective}, \ref{eq:constraint}). We select the data of the last day in the dataset to carry out the optimization experiment. Then, we recount the last day's flow after optimizing passengers' travel routes and calculate the standard deviation of the flow of all bus stops as a simple evaluation metric. From \figurename{ \ref{fig:optimization_route}}, our route optimization has little impact on passenger flow because of our constraint conditions. On the other hand, it reduces $std$, which means balancing the flow of all bus stops, reducing the probability of congestion on the bus, and achieving traffic diversion in bus travel. Hence, the prediction results of our MPGCN are accurate enough to be effectively applied in traffic flow-based downstream tasks.

\section{Discussion}
\par Overall, our studies focus on solving the problem of passenger traffic prediction using a novel concept called passenger mobility pattern. Deep clustering with GCN, according to the result analysis, can implicitly extract passenger mobility patterns in the sharing-stop network. In terms of spatial characteristics, passengers of the same mobility pattern have high similarities. That is to say, they share similar travel bus stops and routes, and their contributions to specific routes' flow are dominant. However, our established sharing-stop network does not include the relationship of passenger travel time, so we were unable to uncover any potential temporal-related laws in the extracted mobility patterns, which merits more investigation in future. Besides, our studies can also be used in other modes of transport like the subway. However, it should be highlighted that we are primarily interested in mobility patterns based on frequent and consistent travel, which necessitates sufficient co-occurrence relationship for passengers. Therefore, our method may not be suitable for the traffic flow prediction task of infrequently used transport mode. For example, if a passenger only travels by air once or twice a year, it is hard to discover their mobility patterns.

\par In addition, the route optimization problem in public transport systems involves a variety of specific tasks on different scenarios, including route network design, frequency setting, timetable optimization~\cite{borndorfer2017passenger}, schedule optimization~\cite{wu2020predicting}, and passenger assignment adjustment. Travel time, waiting time, path length, amount of crowding on the buses, and other objective functions are all varied due to different sub-problems.~\cite{duran2021survey}. In essence, our case study of route optimization is about passenger assignment adjustment, thus we do not operate on route design, bus timetables, or schedules. Unlike mentioned above, our optimization aims to passenger diversion and lower level of bus crowding, as stated in Section \ref{route_optimization}. It is also understandable that passengers can be informed of route congestion and recommended an alternate travel route to alleviate the overcrowded at bus stops. Furthermore, there are also shortcomings in our route optimization. For example, the new route selected in the optimization may be longer than the original route, resulting in longer travel time or more transfers times. Therefore, our case study is merely a naive application example, which can demonstrate the advantages of passengers prediction in our MPGCN, but it also provides researchers with another idea to extend more applications in practical industrial application.

\section{Conclusion and Future Work}
\par In this paper, we introduced forward a passenger flow prediction framework, MPGCN, including the sharing-stop network construction, passenger mobility patterns recognition, and passenger flow prediction. We executed experiments to analyze extracted mobility patterns, and we discovered that different mobility patterns can fit heavy-tailed distributions well and have their own travel laws and route preference. Our framework gave full consideration to the impact of passenger mobility patterns on prediction. We conducted extensive experiments to demonstrate that MPGCN can accurately predict bus passenger flow based on the real bus record data. Finally, we design a simple case study, which shows the value and application of our accurate prediction in the downstream tasks, like route optimization. Besides, prediction results of MPGCN can be applied extensively in other services and strategies of ITS for sustainable public transportation, such as subsequent bus scheduling, route planning, and congestion management.

\par When sufficient multi-source sensors data is available, we will attempt to provide fine-grained analysis and service based on human mobility patterns. For further research, considering specific Spatio-temporal information, we will construct different personal mobility prediction architectures.

\appendix[Analysis of Passenger Mobility Patterns]
\par \tablename{ \ref{tbl:patterns_proportion_routes}} shows flow contribution proportion of passengers of four mobility patterns in the top 40 bus routes by total passenger flow, which is used to analyze the route preferences of mobility patterns and travel rules.

\begin{table*}
	\caption{Distribution proportion of passengers of four mobility patterns}
	\label{tbl:patterns_proportion_routes}
	\centering
	\begin{tabular}{|l|l|l|l|l|l|l|l|l|l|l|l|}
		\hline
		\multicolumn{1}{|c|}{\multirow{2}{*}{Route No.}} & \multicolumn{4}{c|}{Pattern No. (\%)}                                                             & \multicolumn{1}{c|}{\multirow{2}{*}{Total flow}} & \multicolumn{1}{c|}{\multirow{2}{*}{Route No.}} & \multicolumn{4}{c|}{Pattern No. (\%)}                                                             & \multicolumn{1}{c|}{\multirow{2}{*}{Total flow}} \\ \cline{2-5} \cline{8-11}
		\multicolumn{1}{|c|}{}                           & \multicolumn{1}{c|}{1} & \multicolumn{1}{c|}{2} & \multicolumn{1}{c|}{3} & \multicolumn{1}{c|}{4} & \multicolumn{1}{c|}{}                            & \multicolumn{1}{c|}{}                           & \multicolumn{1}{c|}{1} & \multicolumn{1}{c|}{2} & \multicolumn{1}{c|}{3} & \multicolumn{1}{c|}{4} & \multicolumn{1}{c|}{}                            \\ \hline
		1                                               & 31.38                  & 65.48                  & 1.61                   & 1.53                   & 51736                                            & 32                                             & 67.47                  & 13.67                  & 18.37                  & 0.49                   & 15335                                            \\ \hline
		28                                              & 28.57                  & 62.74                  & 7.18                   & 1.51                   & 31822                                            & 89                                             & 95.62                  & 2.98                   & 1.09                   & 0.31                   & 15278                                            \\ \hline
		16                                              & 47.9                   & 49.08                  & 2.13                   & 0.89                   & 28659                                            & 33                                             & 64.29                  & 16.68                  & 17.83                  & 1.2                    & 14689                                            \\ \hline
		3                                               & 42.37                  & 18.18                  & 39.06                  & 0.4                    & 26413                                            & 53                                             & 62.31                  & 17.14                  & 20.14                  & 0.41                   & 14543                                            \\ \hline
		4                                               & 44.77                  & 30.65                  & 18.6                   & 5.98                   & 24472                                            & 38                                             & 20.38                  & 70.68                  & 6.59                   & 2.35                   & 14354                                            \\ \hline
		10                                              & 2.95                   & 32.7                   & 1.63                   & 62.72                  & 23871                                            & 50                                             & 18.98                  &17.23                  & 58.64                  & 5.16                   & 14233                                            \\ \hline
		12                                              & 74.03                  & 21.6                   & 2.52                   & 1.84                   & 23593                                            & 63                                             & 0.3                    & 1.74                   & 0.47                   & 97.49                  & 14167                                            \\ \hline
		69                                              & 22.57                  & 12.81                  & 10.32                  & 54.3                   & 22987                                            & 14                                             & 4.67                   & 85.88                  & 2.11                   & 7.34                   & 12390                                            \\ \hline
		11                                              & 28.1                   & 64.2                   & 6.32                   & 1.38                   & 22418                                            & 88                                             & 64.2                   & 14.78                  & 17.6                   & 3.43                   & 11007                                            \\ \hline
		26                                              & 36.95                  & 45.66                  & 12.87                  & 4.52                   & 21992                                            & 91                                             & 19.85                  & 19.45                  & 59.51                  & 1.18                   & 10900                                            \\ \hline
		2                                               & 37.83                  & 45.26                  & 16.47                  & 0.45                   & 21705                                            & 6                                              & 24.9                   & 72.96                  & 0.99                   & 1.16                   & 10893                                            \\ \hline
		23                                              & 20.23                  & 65.06                  & 13.94                  & 0.77                   & 21225                                            & 65                                             & 1.64                   & 5.62                   & 2.32                   & 90.43                  & 10884                                            \\ \hline
		18                                              & 66.34                  & 14.83                  & 17.81                  & 1.02                   & 21199                                            & 5                                              & 73.25                  & 19.92                  & 5.45                   & 1.38                   & 10689                                            \\ \hline
		116                                             & 44.96                  & 52.65                  & 1.61                   & 0.79                   & 21103                                            & 100                                            & 81.18                  & 14.96                  & 2.98                   & 0.89                   & 10504                                            \\ \hline
		31                                              & 42.51                  & 46.65                  & 9.57                   & 1.26                   & 18991                                            & 39                                             & 87.42                  & 9.75                   & 1.33                   & 1.51                   & 10291                                            \\ \hline
		22                                              & 35.62                  & 61.89                  & 1.84                   & 0.65                   & 18792                                            & 66                                             & 0.56                   & 1.39                   & 0.28                   & 97.76                  & 10192                                            \\ \hline
		7                                               & 68.24                  & 27.26                  & 4.17                   & 0.33                   & 17846                                            & 20                                             & 1.62                   & 15.48                  & 0.47                   & 82.44                  & 9872                                             \\ \hline
		62                                              & 1.52                   & 2.1                    & 1.13                   & 95.24                  & 17319                                            & 713                                            & 18.21                  & 19.56                  & 60.81                  & 1.42                   & 9664                                             \\ \hline
		36                                              & 79.62                  & 16.33                  & 3.59                   & 0.46                   & 17060                                            & K1                                              & 37.68                  & 58.37                  & 1.6                    & 2.34                   & 9660                                             \\ \hline
		15                                              & 76.89                  & 17.74                  & 4.97                   & 0.4                    & 16705                                            & 19                                             & 94.81                  & 3.54                   & 0.94                   & 0.71                   & 9633                                             \\ \hline
	\end{tabular}
\end{table*}

\bibliographystyle{IEEEtran}

\bibliography{references}

\vspace{-10mm} 

\begin{IEEEbiography}[{\includegraphics[width=1in,height=1.25in,clip,keepaspectratio]{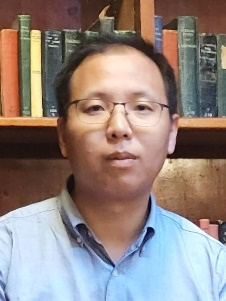}}]{Xiangjie Kong}
	(M’13–SM’17) received the B.Sc. and Ph.D. degrees from Zhejiang University, Hangzhou, China. He is currently a Full Professor with College of Computer Science and Technology, Zhejiang University of Technology. Previously, he was an Associate Professor with the School of Software, Dalian University of Technology, China. He has published over 160 scientific papers in international journals and conferences (with over 130 indexed by ISI SCIE). His research interests include network science, mobile computing, and computational social science.
\end{IEEEbiography}

\vspace{-10mm} 

\begin{IEEEbiography}[{\includegraphics[width=1in,height=1.25in,clip,keepaspectratio]{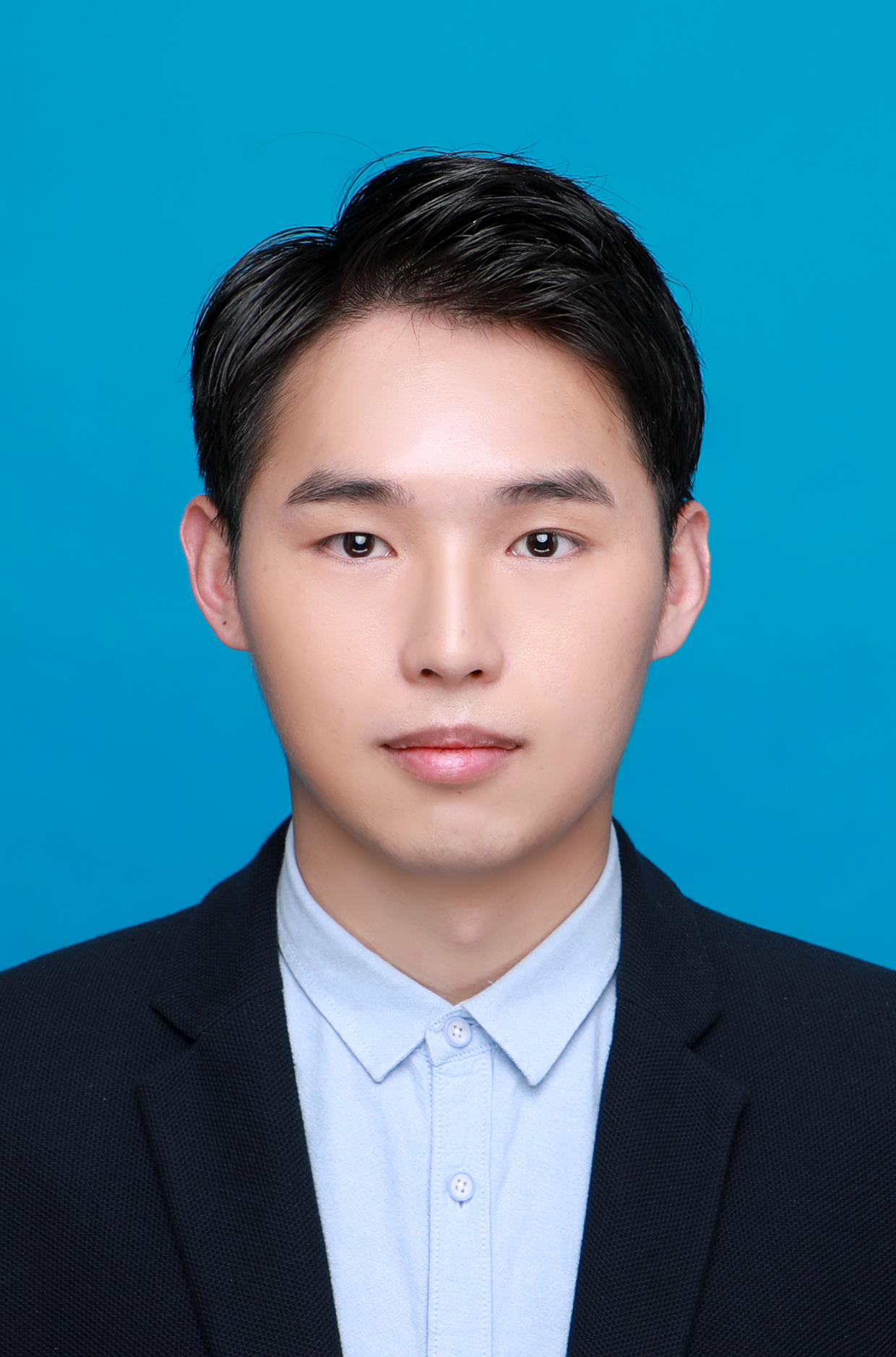}}]{Kailai Wang}
	received the B.Sc. degree in software engineering from the Dalian University of Technology, China, in 2019, where he is currently pursuing the master’s degree with the School of Software. His research interests include analysis of complex networks, network science, and urban computing.
\end{IEEEbiography}

\vspace{-10mm} 

\begin{IEEEbiography}[{\includegraphics[width=1in,height=1.25in,clip,keepaspectratio]{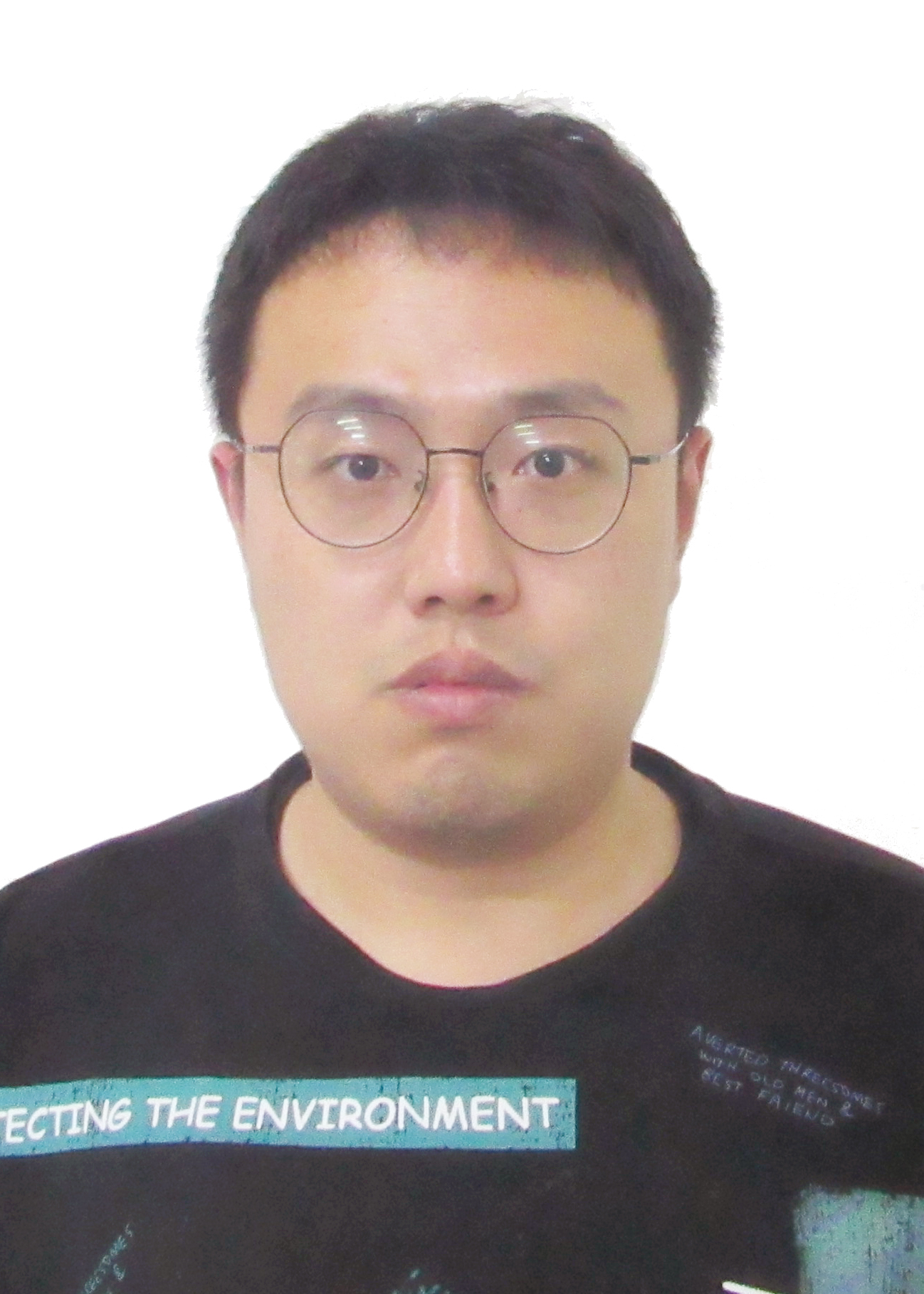}}]{Mingliang Hou}
	received the B.Sc. degree from Dezhou University and the M.Sc. degree from Shandong University, Shandong, China. He is currently pursuing the Ph.D. degree in software engineering with the Dalian University of Technology, Dalian, China. His research interests include graph learning, city science and social computing.
\end{IEEEbiography}

\vspace{-10mm} 

\begin{IEEEbiography}[{\includegraphics[width=1in,height=1.25in,clip,keepaspectratio]{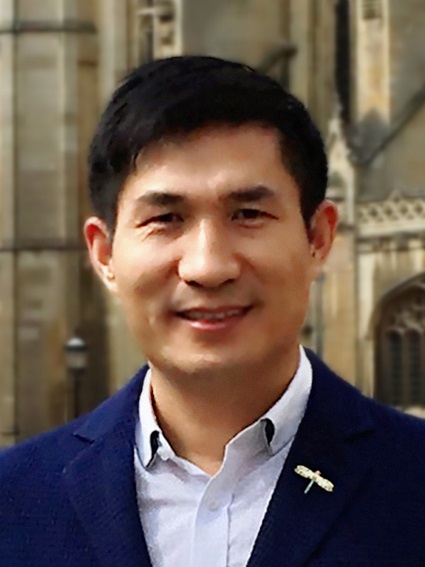}}]{Xia Feng}
	(M’07-SM’12) received the BSc and PhD degrees from Zhejiang University, Hangzhou, China. He is currently an Associate Professor and Discipline Leader in School of Engineering, IT and Physical Sciences, Federation University Australia. Dr. Xia has published 2 books and over 300 scientific papers in international journals and conferences. His research interests include data science, social computing, and systems engineering. He is a Senior Member of IEEE and ACM.
\end{IEEEbiography}

\vspace{-10mm} 

\begin{IEEEbiography}[{\includegraphics[width=1in,height=1.25in,clip,keepaspectratio]{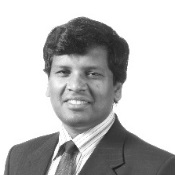}}]{Gour C. Karmakar}
	(M’01) received a B.Sc. degree in Computer Science Engineering from CSE, BUET in 1993 and Masters and Ph.D. degrees in Information Technology from the Faculty of Information Technology, Monash University, in 1999 and 2003, respectively. He is currently an Associate Professor at Federation University Australia. He has published over 161 peer-reviewed research publications including 39 international peer-reviewed reputed journal papers and was awarded six best papers in reputed international conferences. He received a prestigious ARC linkage grant in 2011. His research interest includes multimedia signal processing, traffic signal management, big data analytics, Internet of Things and cybersecurity including trustworthiness measure.
\end{IEEEbiography}

\vspace{-10mm} 

\begin{IEEEbiography}[{\includegraphics[width=1in,height=1.25in,clip,keepaspectratio]{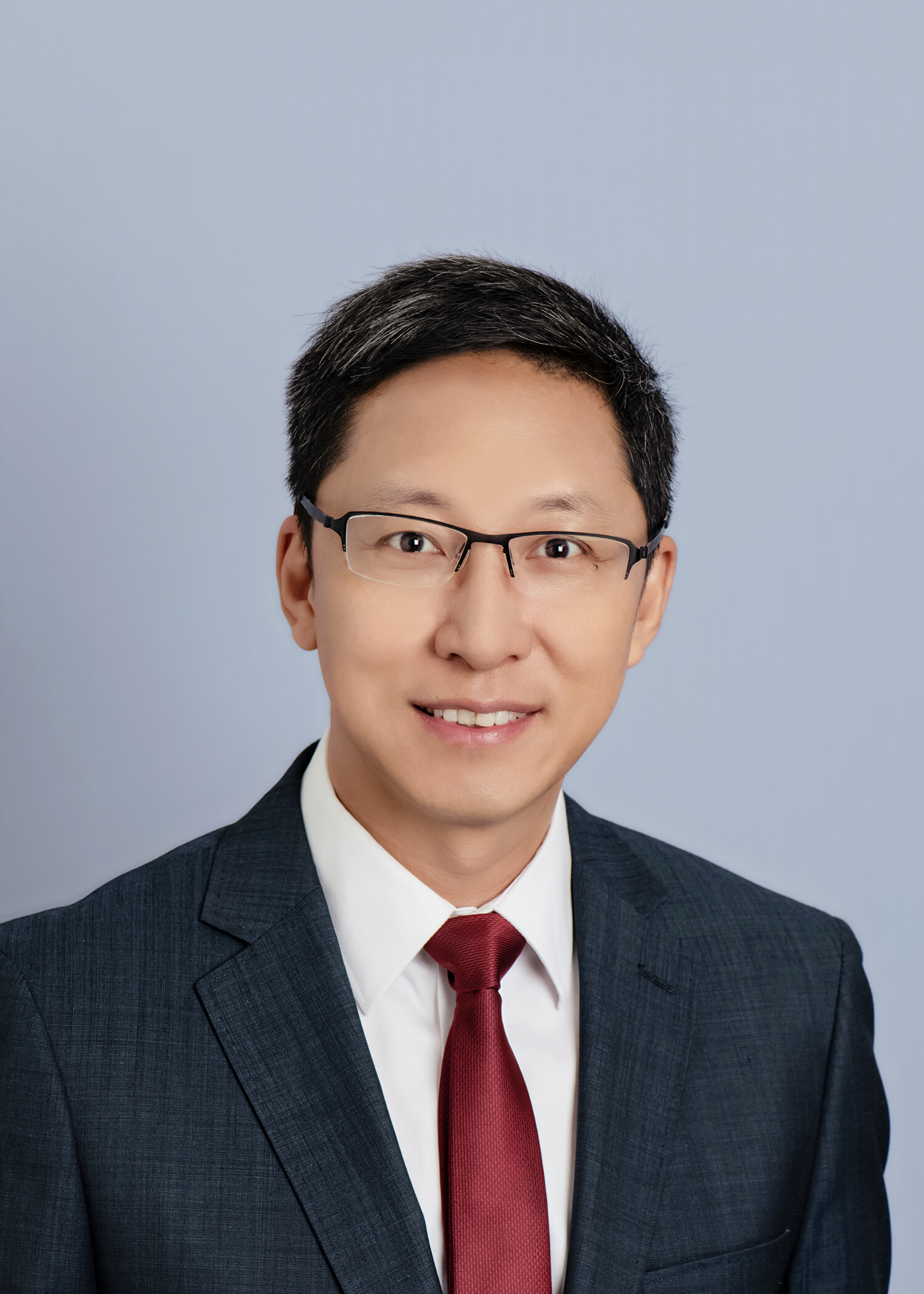}}]{Jianxin Li} 
	received the Ph.D. degree in computer science from Swinburne University of Technology, Australia, in 2009. He is an Associate Professor of Data Science in the School of Information Technology, Deakin University. He has published 90 high quality paper in top-tier venues, including The VLDB Journal, IEEE TKDE, PVLDB and IEEE ICDE. He has received two competitive grants from Australian Research Council. His research interests include graph query processing, social network computing, and information network data analytics.
	
\end{IEEEbiography}

\vspace{-10mm} 

\end{document}